\documentclass[10pt]{article} 

\PassOptionsToPackage{dvipsnames}{xcolor}
\usepackage[accepted]{rlj}

\usepackage{amssymb}            
\usepackage{mathtools}          
\usepackage{mathrsfs}           
\usepackage{graphicx}           
\usepackage{subcaption}         
\usepackage[space]{grffile}     
\usepackage{url}                
\usepackage{lipsum}             
\usepackage{times}
\usepackage{soul}
\usepackage{url}
\usepackage[utf8]{inputenc}
\usepackage{amsmath}
\usepackage{amsthm}
\usepackage{booktabs}
\usepackage{algorithm}
\usepackage[noend]{algpseudocode}
\usepackage{comment}

\usepackage{amsfonts}
\usepackage{pdfrender}
\newcommand*{\boldcheckmark}{%
  \textpdfrender{
    TextRenderingMode=FillStroke,
    LineWidth=.6pt, 
  }{\checkmark}%
}

\newcommand{\rlsquare}{RL\kern-1pt$^2$ }
\newcommand{\rlcube}{RL\kern-1pt$^3$ }
\newcommand{\rlsquarex}{RL\kern-1pt$^2$}
\newcommand{\rlcubex}{RL\kern-1pt$^3$}
\newcommand{\greencheck}{\textcolor{ForestGreen}{\boldcheckmark}}
\newcommand{\orangecheck}{\textcolor{Apricot}{\boldcheckmark}}
\newcommand{\redcross}{\textcolor{BrickRed}{\text{\sffamily x}}}

\title{\rlcubex: Boosting Meta Reinforcement Learning \\ via RL inside \rlsquare}

\setrunningtitle{\rlcubex: Boosting Meta Reinforcement Learning via RL inside \rlsquare}

\author{Abhinav Bhatia\textsuperscript{1}, Samer B. Nashed\textsuperscript{1,2,3}, Shlomo Zilberstein\textsuperscript{1}}

\emails{abhinavbhati@cs.umass.edu, samer.nashed@mila.quebec, shlomo@cs.umass.edu}

\affiliations{
$^{1}$\textbf{Manning College of Information and Computer Sciences, University of Massachusetts Amherst}\\
$^{2}$\textbf{Université de Montréal}\\
$^{3}$\textbf{Mila - Quebec Artificial Intelligence Institute}
}

\contribution{
    A thorough investigation of the hypothesis that augmenting Meta-RL inputs with task specific $Q$-value estimates improves performance across several metrics, which to some may be surprising as this information is already latent within the original input sequence.
    }
    {
    Although some results we present are strong, \emph{this paper is not attempting to present a state-of-the-art Meta-RL system across the board}. We are interested in determining the effects of augmenting the typical Meta-RL inputs of sequence models solving the `outer-loop' using deep RL algorithms with task specific $Q$-value estimates, without using other privileged information or extra resources which in practice often increase performance significantly and are in theory \emph{compatible} with this work.
    }

\contribution{
    This paper presents thorough theoretical, empirical, and logical arguments for the effectiveness of $Q$-estimate state-augmentation; significant attention is devoted to understanding, from different perspectives, why this method achieves the results we see empirically.
    }
    {
    Previous papers have at times highlighted the importance of object-level value estimation for successful Meta-RL, though never in the form of an algorithm such as \rlcubex. This paper presents a unique method, set of experiments, and additional analysis complementing existing research in our effort to better understand the capabilities and properties of Meta-RL systems that learn to solve the meta-level decision process via deep RL algorithms with sequence models (e.g. recurrent networks, transformers).
    }


\summary{Meta reinforcement learning (Meta-RL) methods such as \rlsquare have emerged as promising approaches for learning data-efficient RL algorithms tailored to a given task distribution. However, they show poor asymptotic performance and struggle with out-of-distribution tasks because they rely on sequence models, such as recurrent neural networks or transformers, to process experiences rather than summarize them using general-purpose RL components such as value functions. In contrast, traditional RL algorithms are data-inefficient as they do not use domain knowledge, but do converge to an optimal policy in the limit. We investigate the hypothesis that incorporating action-values, learned per task via traditional RL, in the inputs to Meta-RL systems that solve the meta-level decision process via an `outer-loop' deep RL algorithm and a sequence model (e.g. recurrent network, transformer) has a positive effect on the above shortcomings. Using an example implementation, called \rlcubex, we demonstrate that this strategy earns greater cumulative reward in the long term compared to \rlsquare while drastically reducing meta-training time and generalizing better to out-of-distribution tasks. Experiments are conducted on both custom and benchmark discrete domains from the Meta-RL literature that exhibit a range of short-term, long-term, and complex dependencies.
}

\begin{document}

\makeCover  
\maketitle  

\begin{abstract}
Meta reinforcement learning (Meta-RL) methods such as \rlsquare have emerged as promising approaches for learning data-efficient RL algorithms tailored to a given task distribution. However, they show poor asymptotic performance and struggle with out-of-distribution tasks because they rely on sequence models, such as recurrent neural networks or transformers, to process experiences rather than summarize them using general-purpose RL components such as value functions. In contrast, traditional RL algorithms are data-inefficient as they do not use domain knowledge, but do converge to an optimal policy in the limit. We propose \rlcubex, a principled hybrid approach that incorporates action-values, learned per task via traditional RL, in the inputs to Meta-RL. We show that \rlcube earns a greater cumulative reward in the long term compared to \rlsquare while drastically reducing meta-training time and generalizes better to out-of-distribution tasks. Experiments are conducted on Meta-RL benchmarks and custom discrete domains that exhibit a range of short-term, long-term, and complex dependencies.\looseness-1
\end{abstract}

\section{Introduction}

Reinforcement learning (RL) has been shown to produce effective policies in a variety of applications, including both virtual~\citep{mnih2015human} and embodied~\citep{schulman2017proximal,haarnoja2018soft} systems. However, traditional RL algorithms have three major drawbacks: they can be slow to converge, require a large amount of data, and often have difficulty generalizing to out-of-distribution (OOD) tasks not practiced during training. These shortcomings are especially glaring in settings where the goal is to learn policies for a collection or distribution of problems that share some similarities, and for which traditional RL must start from scratch for each problem. For example, many robotic manipulation tasks require interacting with an array of objects with similar but not identical shapes, sizes, weights, materials, and appearances, such as mugs and cups. It is likely that effective manipulation strategies for these tasks will be similar, but they may also differ in ways that make it challenging to learn a single policy that is highly successful on all instances. Recently, meta-reinforcement learning—often dubbed ``learning to learn''— has been proposed as an approach to mitigate these shortcomings by deriving RL algorithms (or Meta-RL policies) that \emph{adapt} efficiently to a distribution of tasks that share some common structure \citep{duan2016rl,wang2016learning}.

\begin{table}[b]
    \caption{\rlcube combines the strengths of \rlsquare and traditional RL. Like \rlsquarex, \rlcube uses finite-context sequence models to represent data-efficient RL algorithms, optimized for tasks within a specified distribution. However, \rlcube also includes a general-purpose RL routine that distills arbitrary amounts of data into optimal value-function estimates during adaptation. This improves long-term reasoning and OOD generalization.}
    \centering
    \begin{tabular}{l c c c}
     & \textbf{RL} & \textbf{\rlsquare} & \textbf{\rlcube}\\
    \hline \\
    Short-Term Efficiency & \redcross & \greencheck & \greencheck \\
    Long-Term Performance & \greencheck & \redcross & \greencheck \\
    OOD Generalization & \greencheck & \redcross & \orangecheck \\
     & (General Purpose) &  & (Improved) \\
    \end{tabular}
    \label{tab:rl3strengths}
\end{table}

While Meta-RL systems represent a significant improvement over traditional RL in such settings, they still require large amounts of data during meta-training time, can have poor asymptotic performance during adaptation, and may generalize poorly to tasks not represented in the meta-training distribution. This is partly because they rely on black-box sequence models like recurrent neural networks or transformers to process experience data. These models cannot handle arbitrary amounts of data effectively and lack integrated general-purpose RL components that could induce a broader generalization bias.

Hence, we propose \rlcubex, an approach that embeds the strengths of traditional RL within Meta-RL. Table\;\ref{tab:rl3strengths} highlights our primary aims and the foremost insight informing our approach. The key idea in \rlcube is an additional `object-level' RL procedure executed within the Meta-RL architecture that computes task-specific optimal $Q$-value estimates as supplementary inputs to the meta-learner, in conjunction with sequences of states, actions and rewards. In principle, our approach allows the meta-learner to learn how to optimally fuse raw experience data with summarizations provided by the $Q$-estimates. Ultimately, \rlcube leverages $Q$-estimates' generality, ability to compress large amounts of experiences into useful summaries, direct actionability, and asymptotic optimality to enhance long-term performance and OOD generalization and drastically reduce meta-training time.

While $Q$-value estimates can be injected into any other Meta-RL algorithm, for clarity of exposition, we implement \rlcube by injecting $Q$-value estimates into one of the most popular and easily understood Meta-RL algorithm, \rlsquare \citep{duan2016rl}, hence, the name \rlcubex. However, it should be noted that \emph{our baseline implementation of \rlsquare includes significant enhancements} like using transformers instead of LSTMs to improve long-context reasoning, in addition to incorporating numerous recommendations from \citet{pmlr-v162-ni22a} that have been shown to make even recurrent model-free RL algorithms like \rlsquare competitive with strong Meta-RL baselines like VariBAD~\citep{zintgraf2020varibad}.\looseness-1

The primary contribution of this paper is a proof-of-concept that injecting $Q$-estimates obtained via traditional object-level RL, alongside experience histories, leads to improved performance, which to some may be surprising as this information is already latent within the original input sequence. Specifically, we observe higher long-term returns and better OOD generalization, while maintaining short-term efficiency in Meta-RL agents like \rlsquare that run an `outer-loop' deep RL algorithm and use sequence models like LSTMs or transformers. We further demonstrate that our approach can also work with an abstract, or coarse, representation of the object-level MDP. We experiment with discrete domains that both reflect the challenges faced by Meta-RL and simultaneously allow transparent analysis of the results. Finally, we examine the key insights that inform our approach and show theoretically that object-level $Q$-values are directly related to the optimal meta-value function.

\section{Related Work}

Although Meta-RL is a fairly new topic of research, the general concept of meta-learning is decades old~\citep{vilalta2002perspective}, which, coupled with a significant number of design decisions for Meta-RL systems, has created a large number of different proposals for how systems ought to best exploit the resources available within their deployment contexts~\citep{beck2023survey}. At a high level, most Meta-RL algorithms can be categorized as either parameterized policy gradient (PPG) models~\citep{finn2017model,li2017meta,sung2017learning,al2017continuous,gupta2018meta,yoon2018bayesian,stadie2018some,vuorio2019multimodal,zintgraf2019fast,raghu2019rapid,kaushik2020fast,ghadirzadeh2021bayesian,mandi2022effectiveness} or black box models~\citep{duan2016rl,heess2015memory,wang2016learning,foerster2017learning,mishra2017simple,humplik2019meta,fakoor2019meta,yan2020multimodal,zintgraf2020varibad,liu2021decoupling,emukpere2021successor,beck2022hypernetworks}. PPG approaches assume that the underlying learning process is best represented as a policy gradient, where the set of parameters that define the underlying algorithm ultimately form a differentiable set of meta-parameters that the Meta-RL system may learn to adjust. The additional structure provided by this assumption, combined with the generality of policy gradient methods, means that typically PPG methods retain greater generalization capabilities on out-of-distribution tasks. However, due to their inherent data requirements, PPG methods are often slower to adapt and initially train.

In this paper we focus on black box models, which represent the meta-learning function as a neural network, often a recurrent neural network (RNN)~\citep{duan2016rl,heess2015memory,wang2016learning,humplik2019meta,fakoor2019meta,yan2020multimodal,zintgraf2020varibad,liu2021decoupling} or a transformer~\citep{mishra2017simple,wang2021alchemy,melo2022transformers}. There are also several hybrid approaches that combine PPG and black box methods, either during meta-training~\citep{ren2023leveraging} or fine-tuning~\citep{lan2019meta,xiong2021practical}. Using black box models simplifies the process of augmenting meta states with $Q$-estimates and allows us to retain relatively better data efficiency while relying on the $Q$-value injections for better long-term performance and generalization.\looseness-1

Meta-RL systems may also leverage extra information available during training, such as task identification~\citep{humplik2019meta,liu2021decoupling}. Such `privileged information' can of course lead to more performant systems, but is not universally available. As our hypothesis does not rely on the availability of such information, we expect our approach to be orthogonal to, and compatible with, such methods. Black box Meta-RL systems that do not use privileged information still vary in several ways, including the choice between on-policy and off-policy learning and, in systems that use neural networks, the choice between transformers~\citep{vaswani2017attention} and RNNs~\citep{elman1990finding,hochreiter1997long,cho2014learning}.

The most relevant methods to our work are end-to-end methods, which use a single function approximator to subsume both learner and meta-learner, such as \rlsquare \citep{duan2016rl}, L2L \citep{wang2016learning}, SNAIL \citep{mishra2017simple}, and E-\rlsquare \citep{stadie2018some}, and methods that exploit the formal description of the Meta-RL problem as a POMDP or a Bayes-adaptive MDP (BAMDP) \citep{duff2002optimal}. These methods attempt to learn policies conditioned on the BAMDP belief state while also approximating this belief state by, for example, variational inference (VariBAD) \citep{zintgraf2020varibad,dorfman2020offline}, or random network distillation on belief states (HyperX) \citep{zintgraf2021exploration}. Or, they simply encode enough experience history to approximate POMDP beliefs (\rlsquarex) \citep{duan2016rl,wang2016learning}.

Our proposed method is an end-to-end system that exploits the BAMDP structure of the Meta-RL problem by spending a small amount of extra computation to provide inputs to the end-to-end learner that more closely resemble important constituents of BAMDP value functions. Thus, the primary difference between this work and previous work is the injection of $Q$-value estimates into the Meta-RL agent state at each meta-step, in addition to the state-action-reward histories. In this work, our approach, \rlcubex, is implemented by simply injecting $Q$-value estimates into \rlsquare alongside experience history, although any other Meta-RL algorithm can be used.

\section{Background and Notation}

In this section, we briefly cover some notation and concepts upon which this paper is built.

\subsection{Partially Observable MDPs}\label{sec:pomdp}

We use the standard notation defining a Markov decision process (MDP) as a tuple $M = \langle S, A, T, R \rangle$, where $S$ is a set of states, $A$ is a set of actions, $T$ is the transition function, and $R$ is the reward function. A \emph{partially observable Markov decision process} (POMDP) extends MDPs to settings with partially observable states. A POMDP is described as a tuple $\langle S, A, T, R, \Omega, O \rangle$, where $S$, $A$, $T$, and $R$ are as in an MDP. In a POMDP, the agent receives observations $\omega \in \Omega$ instead of direct state information, with an observation function $O(\omega | s', a)$ representing the probability of observing $\omega$ after taking action $a$ and transitioning to state $s'$. POMDPs can also be formulated as continuous-state belief MDPs, where a belief state $b \in \Delta^{|\mathcal{S}|}$ is a probability distribution over all possible states. Beliefs are updated online via Bayesian inference: $b'(s'| b, a, \omega) \propto O(\omega | s', a)\, \sum_{s \in S} T(s' | s, a)\, b(s)$ as new observations are received. The policy maps beliefs to actions, $\pi: \Delta^{|\mathcal{S}|} \rightarrow \mathcal{A}$.

\subsection{Reinforcement Learning}

Reinforcement learning (RL) agents learn an optimal policy in an MDP with unknown dynamics using only transition and reward feedback. This is often done by incrementally estimating the optimal action-value function $Q^*(s,a)$~\citep{watkins1992q}, which satisfies the Bellman optimality equation $Q^*(s, a) = \mathbb{E}_{s'} [R(s, a) + \gamma \max_{a' \in A} Q^*(s', a')]$. In large or continuous state spaces, it is common to use deep neural networks to represent the action-value function~\citep{mnih2015human}. We denote the vector of $Q$-estimates for all actions at state $s$ as $Q(s)$, and after $t$ feedback steps as $Q^t(s)$. Q-learning is known to converge asymptotically~\citep{sutton2018reinforcement}, provided each state-action pair is explored sufficiently. As a rough statement, $||Q^t(s) - Q^*(s)||_{\infty}$ is proportional to $\approx \frac{1}{\sqrt{t}}$, with strong results on convergence error available~\citep{szepesvari1997asymptotic,kearns1998finite,even2003learning}. Typically, the \emph{formal objective} in RL is to optimize the value of the final policy, i.e., the cumulative reward per episode, disregarding the cumulative reward missed (or \emph{regret}) during learning due to suboptimal actions taken by the \emph{behavior policy} used for exploration.

\begin{figure}[t]
    \centering
    \includegraphics[width=\linewidth]{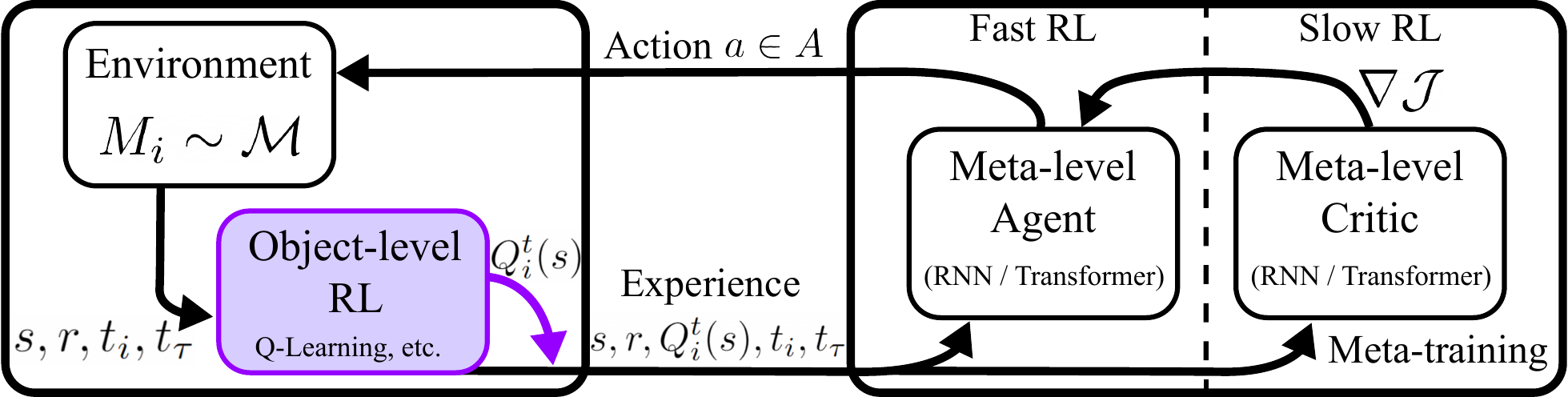}%
    \caption{Overview diagram of \rlcubex. Black entities represent standard components from \rlsquarex, and purple entities represent additions for \rlcubex. $M_i$ is the current MDP; $s$ is a state; $r$ is a reward; $t_i$ and $t_{\tau}$ are the amount of time spent experiencing the current MDP and current episode, respectively; $Q_i^t$ is the $Q$-value estimate for MDP $i$ after $t$ actions; $\nabla \mathcal{J}$ is the policy gradient for meta-training.}
    \label{fig:block_diagram}
\end{figure}

\subsection{Meta Reinforcement Learning}\label{sec:meta-rl}

While standard model-free RL seeks a single optimal policy per MDP, Meta-RL aims to train strategies that rapidly adapt across a distribution of related tasks $\mathcal{M}$. This is especially valuable when the agent faces a fixed interaction budget $H$ per task, assuming shared state and action spaces. Formally, Meta-RL optimizes a `fast' \emph{behavior policy} $\pi$, also called the \emph{Meta-RL policy}, to maximize cumulative reward (or equivalently, minimize regret) during the adaptation period:
\begin{equation}
    \mathcal{J}(\pi) = \mathbb{E}_{M_i \sim \mathcal{M}} \Bigl[ \sum_{t=0}^H \gamma^t\, \mathbb{E}_{(s_t,a_t)\sim\rho_{i,t}^{\pi}}\bigl[R_i(s_t,a_t)\bigr]\Bigr],
\end{equation}
where $\rho_{i,t}^{\pi}$ denotes the state-action distribution induced by $\pi$ in $M_i$ at timestep $t$. Unlike Markovian policies in standard RL, the Meta-RL policy conditions on the \emph{experience history} $\Upsilon = (s_0,a_0,r_0,\dots,s_t)$, which spans multiple episodes within each adaptation period.

Theoretically, Meta-RL can be framed as solving a meta-level POMDP in which the task identity $M_i$ is the hidden variable. Augmenting the state with a belief over tasks $b(i)$ yields a Bayes-Adaptive MDP (BAMDP)~\citep{duff2002optimal,ghavamzadeh2015bayesian}, where the belief state $[s,b]$ captures all necessary information for optimal action selection. In practice, however, many approaches approximate this mapping without explicit belief modeling, instead relying on recurrent or transformer-based policies. A prominent example of this class is RL$^2$~\citep{duan2016rl,wang2016learning}, which we consider representative of Meta-RL methods that solve the meta-level POMDP by using an outer-loop `slow' deep RL algorithm e.g., PPO~\citep{schulman2017proximal} to train a sequence model—such as a recurrent network or transformer—to map experience histories to policies or meta-value functions, thereby optimizing $\mathcal{J}(\pi)$. Our approach builds on this framework by augmenting the experience history with task-specific value estimates, as described below.

\section{RL\texorpdfstring{$^3$}{3}}

To address the limitations of black box Meta-RL methods, we propose \rlcubex, a principled approach that leverages (1) the inherent generality of action-value estimates, (2) their ability to compress experience histories into useful summaries, (3) their direct actionability \& asymptotic optimality, (4) their ability to inform task identification, and (5) their relation to the optimal meta-value function, in order to enhance out-of-distribution (OOD) generalization and performance over extended adaptation periods. The central, novel mechanism in \rlcube is an additional `object-level' RL procedure executed within the Meta-RL architecture, shown in Fig.\;\ref{fig:block_diagram}, that computes task-specific optimal $Q$-value estimates $Q^t_i(s_t)$---denoting $\hat Q^*$ estimates for task $i$ after $t$ transition samples---and state-action counts as supplementary inputs to the Meta-RL policy in conjunction with the sequence of states, actions, and rewards $(s_0, a_0, r_0,..., s_t)$. The $Q$-estimates are computed off-policy and may involve experience replay or model estimation \& planning for greater data efficiency. The estimates and the counts are reset at the beginning of each meta-episode as a new task $M_i$ is sampled. We now present a series of key insights informing our approach.

First, estimating action-values is a key component in many general-purpose, or \textbf{universal}, RL algorithms, and asymptotically, they \emph{fully} inform optimal behavior \emph{irrespective of domain}. Strategies for optimal exploration-exploitation trade-off are domain-dependent and rely on historical data, yet many exploration approaches use estimated $Q$-values and some notion of counts \emph{alone}, such as epsilon-greedy, Boltzmann exploration, upper confidence bounds (UCB/UCT)~\citep{auer2002using,kocsis2006bandit}, count-based exploration~\citep{tang2017exploration}, curiosity-based exploration~\citep{pathak18largescale}, and maximum-entropy RL~\citep{haarnoja2018soft}. This creates a strong empirical case that using $Q$-value estimates and state-action counts for efficient exploration has inherent generality.

\begin{figure}[t]
    \centering
    \includegraphics[width=0.8\linewidth]{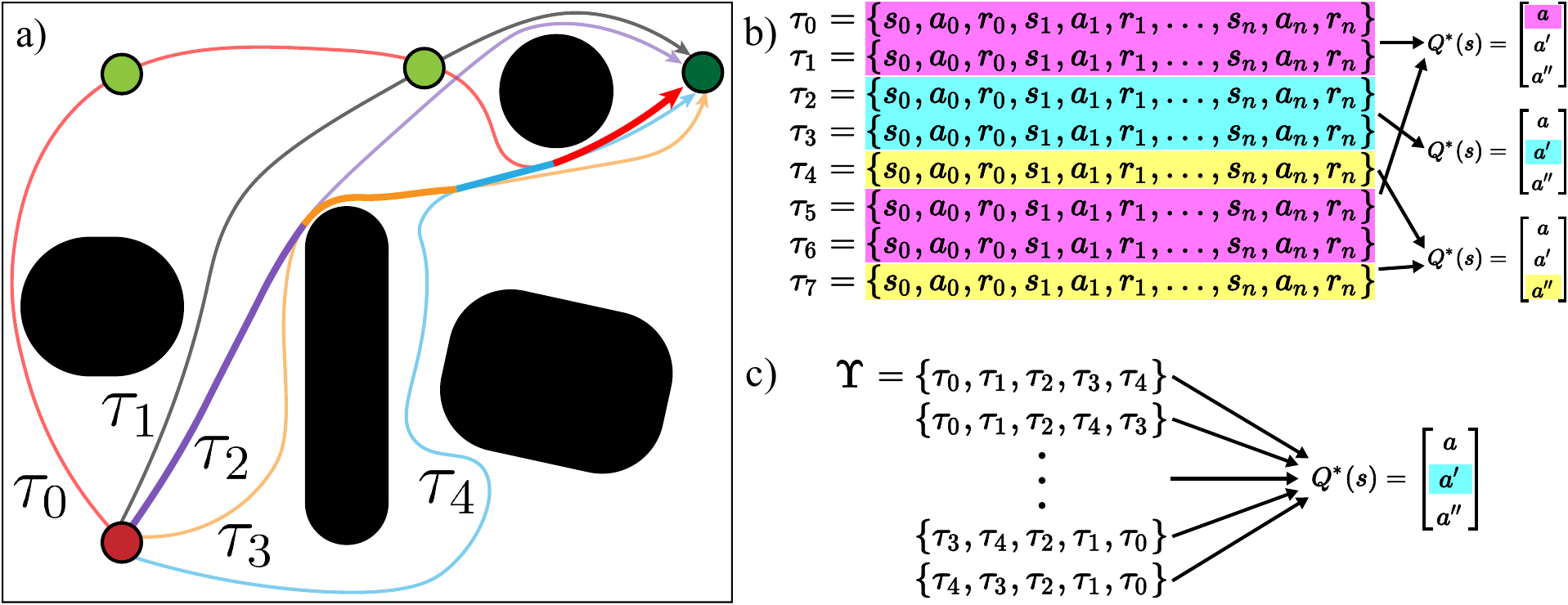}
    \caption{Sub-figure (a) shows a meta-episode in a shortest-path environment where the goal position (green circles) and the obstacles (black regions) may vary across tasks. In this meta-episode, after the Meta-RL agent narrows its belief about the goal position of this task (dark-green circle) having followed a principled exploration strategy ($\tau_0$), it explores potential shorter paths in subsequent episodes ($\tau_1, \tau_2, \tau_3, \tau_4$). Throughout this process, the estimated value-function {$\hat{Q}^*$} implicitly ``remembers" the goal position and previous paths traversed in a finite-size representation, and updates the shortest path calculation (highlighted in bold) using Bellman backups when paths intersect. Sub-figures (b) and (c) illustrate the many-to-one mapping of object- and meta-level data streams to $Q$-estimates, and thus their utility as compression and summarization mechanisms for meta-learning.}
    \label{fig:intuition}
\end{figure}

Second, $Q$-estimates \textbf{summarize experience histories} of arbitrary length \emph{and order} in one constant-size vector. This mapping is many-to-one, and any permutation of transitions ($\langle s,a,r,s' \rangle$ tuples) or episodes in a history of experiences yield the same $Q$-estimates. Although this compression is lossy, it still ``remembers'' important aspects of the experienced episodes, such as high-return actions and goal positions (see Fig.\;\ref{fig:intuition}), since $Q$-estimates persist across episodes. In conjunction, action counts provide an explicit record of exploration coverage. This simplifies the mapping the meta-agent needs to learn, as $Q$-estimates and counts together represent a smaller and more salient set of inputs compared to all possible histories with the same implication.

Third, $Q$-estimates are \textbf{actionable}. Estimated off-policy, they explicitly represent the optimal exploitation policy for the current task given the data, insofar as the RL module is data-efficient, relieving the Meta-RL agent from performing such calculations inside the transformer or RNN. Over time, $Q$-estimates become more reliable and directly indicate the optimal policy, whereas processing raw data becomes more challenging. Provided alongside action counts, they also give the agent an implicit mechanism to modulate trust in $Q$-estimates as evidence accumulates. Fortunately, by incorporating $Q$-estimates, the Meta-RL agent can eventually ignore the history in the long run (or towards the end of the interaction period) and simply exploit the $Q$-estimates by selecting actions greedily.\looseness-1

Fourth, $Q$-estimates are \textbf{excellent task discriminators} and serve as another line of evidence vis-à-vis maintaining belief over tasks. In a simple domain like Bernoulli multi-armed bandits~\citep{duan2016rl}, $Q$-estimates and action counts combined are sufficient for Bayes-optimal behavior even without providing raw experience data—a result surprisingly unstated in the literature to the best of our knowledge (see Appendix~\ref{subsec:Bernoulli-Bandits-proof}). However, $Q$-estimates and counts may not always be sufficient for Bayes-optimal beliefs (see Gaussian multi-armed bandits example in Appendix~\ref{subsec:Gaussian-Bandits-proof}). In more complex domains, it is hard to prove the sufficiency of $Q$-estimates regarding task discrimination. However, via empirical analysis in Appendix~\ref{sec:analysis}, we argue that i) it is highly improbable for two tasks to have similar Q$^*$ functions and ii) $Q$-estimates tend to become accurate task predictors in just a few steps. This implies that the meta-agent may use this finite summary for task inference rather than relying completely on arbitrarily long histories, potentially contributing to enhanced performance over long adaptation periods.

It can be theoretically argued that since the meta-agent is a BAMDP \emph{policy}, it is meta-trained to select greedy actions with respect to the BAMDP meta-value function and thus should not require constructing a task-specific plan internally. However, the optimality of the meta action-value function depends on implicitly (or explicitly in some approaches~\citep{humplik2019meta,zintgraf2020varibad,dorfman2020offline,zintgraf2021exploration}) maintaining a Bayes-optimal belief over tasks in the transformer or RNN architecture. This may be challenging if the task distribution is too broad and the function approximator is not powerful enough to integrate experience histories into Bayes-optimal beliefs, or altogether impossible if there is a distribution shift at meta-test time. This latter condition is common in practice and is a frequent target use case for Meta-RL systems. Incorporating task-specific $Q$-estimates gives the agent a simple alternative (even if not Bayes-optimal) line of reasoning to translate experiences into actions. Incorporating $Q$-estimates thus \textbf{reduces susceptibility to distribution shifts}, since the arguments presented in this section are domain independent.

Finally, $Q$-estimates often converge far more quickly than the theoretical rate of $1/\sqrt{t}$, allowing them to be useful in the short and medium term, since i) most real-world domains contain significant determinism, ii) it is not necessary to estimate $Q$-values for states unreachable by the optimal policy, and iii) optimal Meta-RL policies may represent active exploration strategies in which $Q$-estimates converge faster, or evolve in a manner leading to quicker task identification. This is intuitively apparent in shortest-path problems, as illustrated in Fig.\;\ref{fig:intuition}(a). In a deep neural network, it is difficult to know exactly how $Q$-estimates will combine with state-action-reward histories when approximating the meta-value function. However, as we show below, we can write an equation for the meta-value function in terms of these constituent streams of information, which may explain why this function is seemingly relatively easy to learn compared to predicting meta-values from histories alone.

\subsection{Theoretical Justification}\label{sec:theory}

Throughout this section, without loss of generality, we assume: i) an infinite-horizon setting, ii) that $\mathcal{M}$ denotes the support set of possible MDPs over which a distribution is defined, iii) all MDPs share the same state and action spaces, and iv) rewards are uniformly bounded by $|R_i(s,a)| \le R_{\max}$. We consider the interpretation of Meta-RL as performing RL on a partially observable Markov decision process (POMDP) in which the partially observable factor is the identity of the object-level MDP. We denote meta-level entities (i.e., those belonging to the POMDP) with an overbar; for example, the meta-level value function $\bar{V}$ and the meta-level belief $\bar{b}$. First, we establish that the optimal meta-level value function is upper bounded by the object-level $Q$-values.

\textbf{\emph{Proof.}} For a given state $s$, there exists a maximum object-level optimal value function $V^*_{\max}(s)$ corresponding to some MDP $M_{\max} \in \mathcal{M}$, such that for all MDPs $M_i \in \mathcal{M}$, $V^*_{\max}(s) \ge V^*_i(s)$. The expected cumulative discounted reward experienced by the agent cannot exceed this most optimistic value, since $\bar{V}^*(\bar{b})$ is a weighted average of the individual value functions $V^{\pi_\theta}(s)$, each of which is upper bounded by $V^*_{\max}(s)$. Thus,
\begin{equation}
    \max_{M_i \in \mathcal{M}} V^*_i(s) \ \ge\ \bar{V}^*(\bar{b})
    \quad \forall s \in S.
    \label{eq:v_ineq}
\end{equation}
Combining the asymptotic accuracy of $Q$-estimates with Equation~\ref{eq:v_ineq} yields
\begin{equation}
    \lim_{t \to \infty} \max_{a \in A, M_i \in \mathcal{M}} Q_i^t(s,a) \geq \bar{V}^*(\bar{b}) \quad \forall s \in S. \qed
\end{equation}
Furthermore, if the meta-level observation $\bar{\omega}$ includes the $Q$-value estimates of the current task $M_i$, it can be shown that as $t \to \infty$, the optimal meta-value function approaches the optimal value function for the current task. Specifically, for any $\epsilon > 0$, there exists $\kappa \in \mathbb{N}$ such that for all $t \ge \kappa$,
\begin{equation}
    \Big | \max_{a \in A} \Big [ Q_i^t(s,a) \Big ] - \bar{V}^*(\bar{b}) \Big | \leq \epsilon \quad \forall s \in S.
    \label{eq:QVequiv}
\end{equation}

Equation~\ref{eq:QVequiv} (see proof in Appendix~\ref{subsec:meta-value-proof}) shows that for $t \ge \kappa$, acting greedily with respect to $Q_i^*$ leads to Bayes-optimal behavior, and explicitly maintaining the Bayes-optimal belief over tasks is no longer required. This implies that the experience history $\Upsilon$ can be ignored at that point.

Moreover, Equation~\ref{eq:QVequiv} implies that for $t < \kappa$,
\begin{align}
     \bar{V}^*(\bar{b}) =  \max_{a \in A} \Big [ Q_i^t(s,a) \Big ] + \varepsilon_i(\Upsilon)
\end{align}
where $\varepsilon_i(\Upsilon)$ denotes the error in $Q$-value estimates as a function of the experience history $\Upsilon$. While this error diminishes as $t \to \infty$, in the short run, a function $f(\Upsilon)$ could be learned to either estimate the error or directly approximate $\bar{V}^*(\bar{b})$. The better performance of \rlcube could be explained by either the error $\varepsilon_i(\Upsilon)$ being simpler to estimate, or the meta-agent’s behavior being more robust to errors in $\varepsilon_i(\Upsilon)$ when $Q$-estimates are supplied directly as inputs, compared to errors arising from approximating $\bar{V}^*(\bar{b})$ without such inputs. Additionally, this approach benefits from the fact that the convergence rate of $Q$-estimates suggests a natural, predictable schedule of shifting reliance from $f(\Upsilon)$ to $Q_i^t(s)$ as $t \to \infty$. However, we do not impose this structure explicitly in the network and instead allow it to learn implicitly how much to rely on the $Q$-estimates, informed by the context and supported by the provided action counts as supplementary signals.

Finally, we note that achieving accurate approximation of $\bar{V}^*(\bar{b})$ as $t \to \infty$ reduces the error in meta-value estimation for all preceding belief states, since meta-values for consecutive beliefs $\bar{b}$ and $\bar{b}'$ are linked via the Bellman equation for BAMDPs (see Appendix~\ref{subsec:meta-value-proof}):
\begin{equation}
    \bar{V}^*(\bar{b}) = \max_{a \in A} \Big [ \sum_{M_i \in \mathcal{M}} \bar{b}(i) R_i(s,a) + \gamma \sum_{\bar{\omega} \in \bar{\Omega}} \Pr(\bar{\omega} |\bar{b},a) \bar{V}^*(\bar{b}') \Big ],
    \label{eq:bamdp}
\end{equation}

This dependency helps meta-training in \rlcube when using temporal-difference learning algorithms. Without conditioning on $Q$-estimates, the error in $\bar{V}^*(\bar{b})$ could instead grow as $t \to \infty$, since the meta-critic must condition on an increasingly large experience history $\Upsilon$. This could destabilize meta-value learning across all preceding belief states during meta-training.

\subsection{Implementation}\label{sec:impl}

Implementing \rlcube involves replacing each MDP in the task distribution with a value-augmented MDP (VAMDP) and solving the resulting distribution using \rlsquarex. Each VAMDP has the same action space and reward function as the original MDP. The value-augmented state $\hat s_t \in S \times \mathbb{R}^k \times \mathbb{I}^k$ comprises the object-level state $s_t$, $k$ real values and $k$ integer values for the $Q$-estimates $Q^t(s_t,a)$ and action counts $N^t(s_t,a)$ for each action. In practice, we represent $Q$-estimates in a numerically stabilized form by decomposing them into their maximum value and centered differences, i.e., providing $\max_a Q^t(s_t,a)$ and the offsets $Q^t(s_t,a)-\max_a Q^t(s_t,a)$, which correspond to the value function and advantages, respectively. This reparametrization is numerically equivalent to raw $Q$-values but improves training stability by normalizing input scale. When $S$ is discrete, $s_t$ is encoded as an $|S|$-dimensional one-hot vector. The value-augmented state space is thus continuous. In the VAMDP transition function, the object-level state $s$ evolves according to the original dynamics, while $Q$-estimate dynamics depend on $T$, $R$, and the specific RL algorithm used. A VAMDP episode spans the full interaction period with the MDP, as $Q$-estimates evolve across multiple episodes. In code, a VAMDP environment is implemented as a wrapper over the base MDP. Pseudocode, additional details, and hyperparameters for \rlsquare and \rlcube are provided in Appendix~\ref{sec:architecture}.

\section{Experiments}

We compare \rlcube to our enhanced implementation of \rlsquarex. In our implementation, we replace LSTMs with transformers in both the meta-actor and meta-critic for the purpose of mapping experiences to actions and meta-values, respectively. This is done to improve \rlsquare's ability to handle long-term dependencies instead of suffering from vanishing gradients. Moreover, \rlsquarex-transformer trains significantly faster than \rlsquarex-LSTM. Second, we include in the state space the total number of interaction steps and the total number of steps within each episode during a meta-episode (see Fig.\;\ref{fig:block_diagram}). Third, we use PPO \citep{schulman2017proximal} for training the meta actor-critic, instead of TRPO~\citep{schulman2015trust}. These modifications and other minor-implementation details incorporate the recommendations made by \citet{pmlr-v162-ni22a}, who show that model-free recurrent RL is competitive with strong Meta-RL approaches such as VariBAD \citep{zintgraf2020varibad}, if implemented properly. \rlcube simply applies the modified version of \rlsquare to the distribution of value-augmented MDPs explained in section \ref{sec:impl}. Within each VAMDP, our choice of object-level RL is a model-based algorithm to maximize data efficiency -- we estimate a tabular model of the environment and run finite-horizon value-iteration using the model. Once again, we emphasize that the core of our approach, which is augmenting MDP states with action-value estimates, is not inherently tied to \rlsquare and is orthogonal to most other Meta-RL research. VAMDPs can be plugged into any base Meta-RL algorithm with a reasonable expectation of improving it.

In our test domains, each meta-episode involves procedurally generating an MDP according to a parameterized distribution, which the meta-actor interacts with for a fixed adaptation period, or interaction budget, $H$. This interaction might consist of multiple object-level episodes of variable length, each of which are no longer than a maximum task horizon. For a given experiment, each approach is trained on the same series of MDPs. Each experiment is done for 3 seeds and the results of the median performing model are reported. For testing, each approach is evaluated on an identical set of 1000 MDPs distinct from the training MDPs. For testing OOD generalization, MDPs are generated from distributions with different parameters than in training. We select three discrete domains for our experiments, which cover a range of short-term, long-term, and complex dependencies. These domains both reflect the challenges faced by Meta-RL and simultaneously allow transparent analysis of the results.

\noindent \textbf{Bernoulli Bandits}: We use the same setup described by~\citet{duan2016rl} with $k=5$ arms. To test OOD generalization, we generate bandit tasks by sampling success probabilities from $\mathcal{N}(0.5, 0.5)$. We should note that this is an easy domain and serves as a sanity check to ensure that $Q$-value estimates do not hurt \rlcubex, causing inferior performance.

\noindent \textbf{Random MDPs:} We use the same setup described by~\citet{duan2016rl}. The MDPs have 10 states, 5 actions, and task horizon 10. The rewards and transition probabilities are drawn from a normal and a flat Dirichlet distribution ($\alpha=1.0$), respectively. OOD test MDPs use Dirichlet $\alpha=0.25$. We should note that this domain is particularly challenging for \rlcube due to the high degree of stochasticity and thus the slower convergence rate of $Q$-estimates.

\begin{table}
    \caption{Test scores (mean $\pm$ standard error) for Bandits domain and the ${}^\dagger$OOD variation.}
    \label{tab:results:bandits}
    \centering
    \begin{tabular}{l c c c}
    \textbf{Budget $H$} & \textbf{\rlsquare} & \textbf{\rlcube} & \textbf{\rlcube (Markov)}\\
    \hline \\
    100 & $76.9 \pm 0.6$ & $77.5 \pm 0.5$ & $75.2 \pm 0.5$ \\
    500 & $392.1 \pm 2.5$ & $393.2 \pm 2.7$ & $391.75 \pm 2.6$ \\
    500$^\dagger$ & $430.2 \pm 2.8$ & $\mathbf{434.9}\pm2.8$ & $433.7 \pm 2.8$\\
    \end{tabular}
\end{table}

\noindent \textbf{GridWorld Navigation:} A set of navigation tasks in a 2D grid environment. We experiment with 11x11 (121 states) and 13x13 (169 states) grids. The agent starts in the center and needs to navigate through obstacles to a single goal. The grid also contains slippery tiles, dangerous tiles and warning tiles. See Fig.\;\ref{fig:gw-example}(a) for an example of a 13x13 grid. The state representation is coordinates $(x,y)$. To test OOD generalization, we vary parameters including the stochasticity of actions, density of obstacles and the number of dangerous tiles. For this domain, we consider an additional variation of \rlcubex, called \rlcubex-\textit{coarse} where a given grid is partitioned into clusters of 2 adjacent tiles (or abstract states), which are used \emph{solely} for the purpose of estimating the object-level $Q$-values. Our goal is to test whether coarse-level $Q$-value estimates are still useful to the Meta-RL policy. The domains and the abstraction strategy are described in greater detail in Appendices \ref{sec:domains} and \ref{sec:rl3appendix}, respectively.

\section{Results}

In summary, we observe that beyond matching or exceeding the performance of \rlsquarex-transformer in all test domains i) \rlcube shows better OOD generalization, which we attribute to the increased generality of the $Q$-value representation, ii) the advantages of \rlcube increase with longer interactions periods and less stochastic tasks, which we attribute to the increased accuracy of the $Q$-value estimates, iii) \rlcube performs well even with coarse-grained object-level RL over abstract states with substantial computational savings, showing minimal drop in performance in most cases, and iv) \rlcube shows faster meta-training.

\begin{figure}[t]
    \centering
    \includegraphics[width=0.29\linewidth]{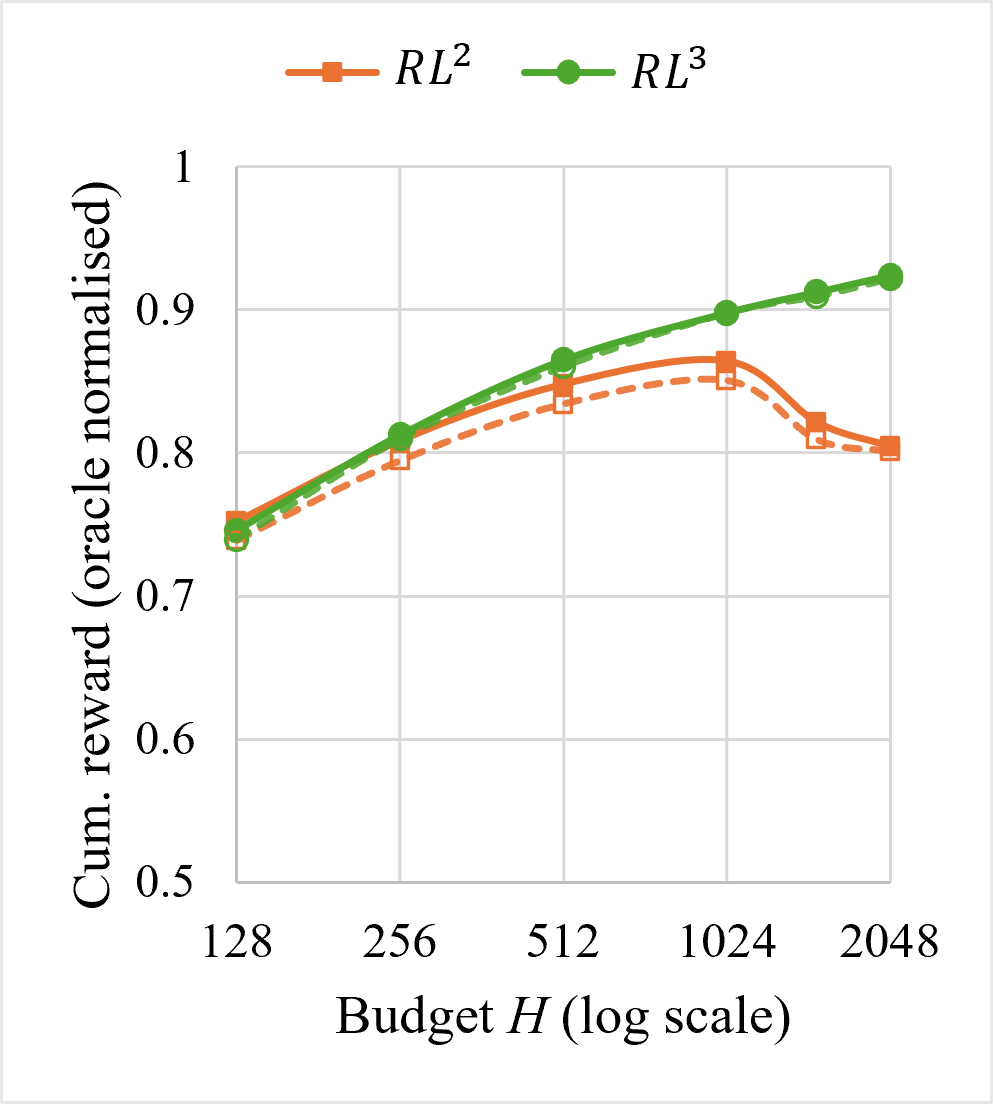}
    \includegraphics[width=0.29\linewidth]{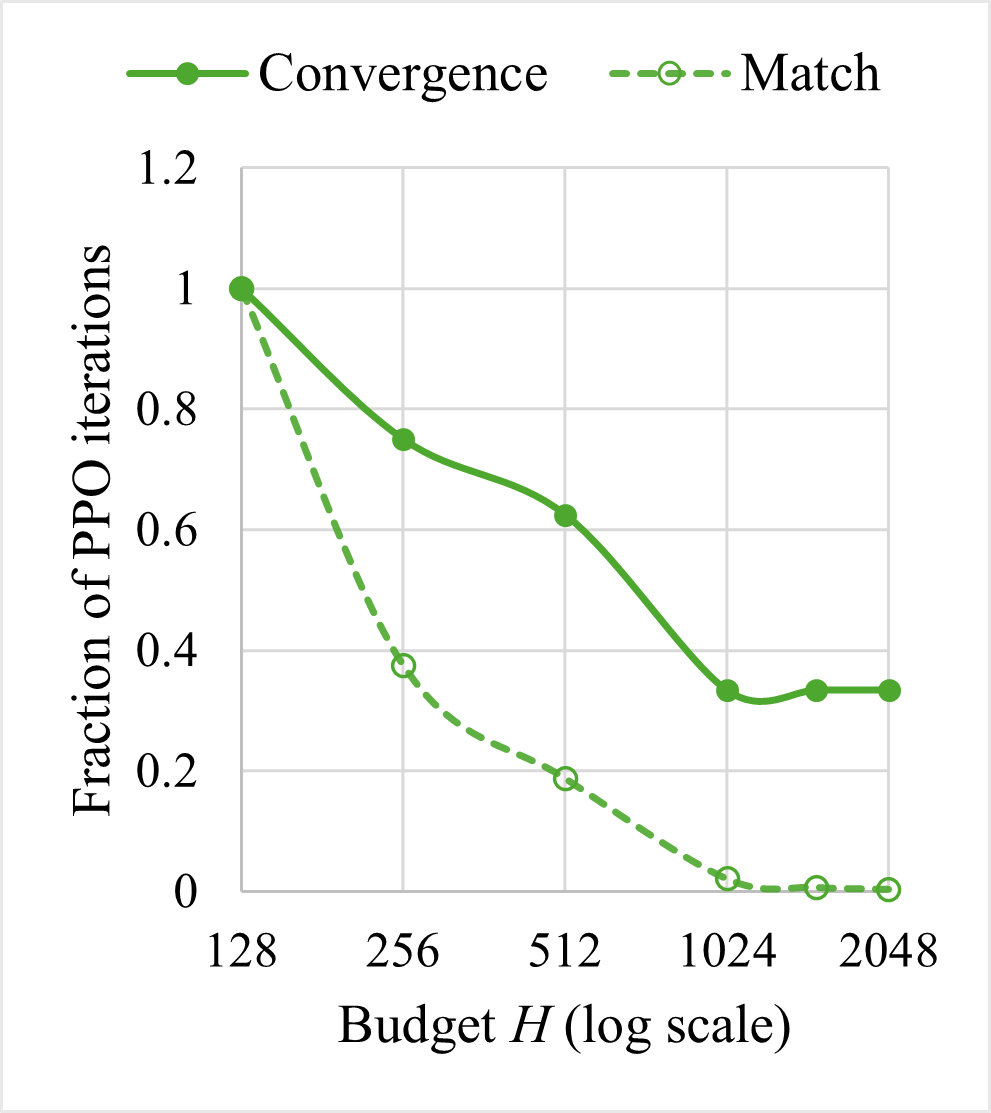}
    \includegraphics[width=0.39\linewidth]{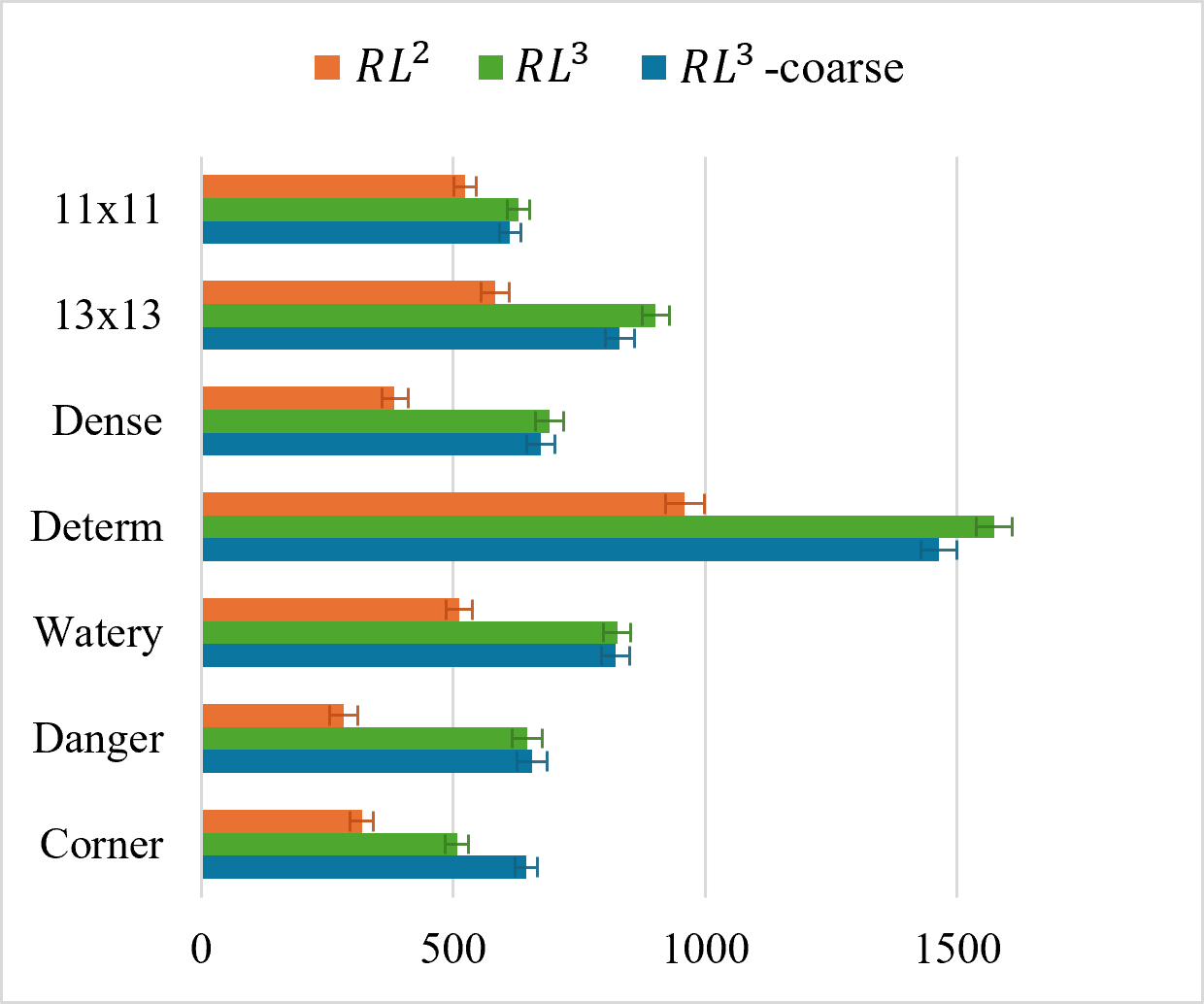}
    \caption{Results for the MDPs and GridWorlds domains. Figure \ref{fig:results:mdps_gw}a shows the average cumulative reward (negligible standard error) earned as a fraction of the oracle policy for in-distribution (solid) and OOD (dashed) tasks; Figure \ref{fig:results:mdps_gw}b shows the fraction of \rlsquarex-transformer meta-training iterations that \rlcube requires (variance is insignificant across seeds) to match \rlsquarex-transformer performance or fully converge, both as functions of the adaptation period $H$. Note the log horizontal axis on both plots. Figure \ref{fig:results:mdps_gw}c shows the average cumulative reward ($\pm$ standard error) earned by \rlsquarex-transformer, \rlcube and \rlcubex-coarse agents on several variations of the GridWorlds domain.}
    \label{fig:results:mdps_gw}
\end{figure}

\noindent \textbf{Bandits:} Fig\;\ref{tab:results:bandits} shows the results for this sanity-check domain. For $H=100$ and $H=500$, both approaches perform comparably. However, the OOD generalization for \rlcube is slightly better. We also experiment with a Markovian version of \rlcubex, where a feed-forward neural network is conditioned only on the $Q$-estimates and action-counts, since those are sufficient for Bayes-optimal behavior in this domain. As expected, the results are similar to regular \rlcube.

\noindent \textbf{MDPs:} Figures\;\ref{fig:results:mdps_gw}(a) and \ref{fig:results:mdps_gw}(b) show the results for the MDPs domain. In Figure\;\ref{fig:results:mdps_gw}(a), we see that for relatively short budgets, $H \leq 500$, both \rlcube and \rlsquarex-transformer perform comparably on in-distribution problems, with \rlcube performing slightly better on OOD tasks. We suspect that, due to the short budgets and highly stochastic domain, $Q$-estimates do not converge enough to be very useful for \rlcube. However, as the budget increases, we see that \rlcube continues to improve while \rlsquarex-transformer actually becomes worse and the performance gap on both in-distribution and OOD tasks becomes significant. Overall, we see that \rlcube \emph{preserves asymptotic scaling properties of traditional RL while simultaneously maintaining strong OOD performance}. Moreover, \rlcube it is able to learn meta-policies much more efficiently. Figure \ref{fig:results:mdps_gw}(b) shows the number of iterations of PPO \rlcube takes to converge completely, as well as to match the performance of \rlsquarex-transformer, measured as a fraction of the time it takes for \rlsquarex-transformer to converge. This advantage of \rlcube is again most pronounced for longer adaptation periods, but we still do observe significant meta-training speedup on even moderate ones. The training curves are available in Appendix \ref{sec:rl3_training_curves}. Overall, it is clear that as adaptation periods grow, \rlcube achieves nearer-to-optimal policies in a fraction of the meta-training time and maintains better OOD generalization.

\noindent \textbf{GridWorlds:} Fig\;\ref{fig:results:mdps_gw}(c) shows the results for the GridWorlds domain. On 11x11 grids with $H=250$, \rlcube significantly outperforms \rlsquarex-transformer. On 13x13 grids with $H=350$, the performance margin is even greater, showing that while \rlsquarex-transformer struggles with a greater number of states, a longer adaptation period and more long-term dependencies, \rlcube can take advantage of the $Q$-estimates to overcome the challenge. We also test the OOD generalization of both approaches in different ways by varying certain parameters of the 13x13 grids, namely, increasing the obstacle density (\textsc{Dense}), making actions on non-water tiles deterministic (\textsc{Deterministic}), increasing the number of wet `W' tiles (\textsc{Watery}), increasing the number of danger `X' tiles (\textsc{Dangerous}) and having the goal only in the corners (\textsc{Corner}). On all variations, \rlcube continues to significantly outperform \rlsquarex-transformer. In a particularly interesting outcome, both approaches show improved performance on the \textsc{Deterministic} variation. However, \rlcube gains 80\% more points than \rlsquarex-transformer, which is likely because $Q$-estimates converge faster on this less stochastic MDP. Conversely, in the \textsc{Watery} variation, which is more stochastic, both \rlsquarex-transformer and \rlcube lose roughly equal number of points. In each case, \rlcubex-coarse significantly outperforms \rlsquarex-transformer. In fact, it performs on par with \rlcubex, even outperforming it on \textsc{Corner} variation, except on the canonical 13x13 case and its \textsc{Deterministic} variation, where it scores about 90\% of the scores for \rlcubex. Finally, we see similar meta-training speedups where \rlcube requires just $50\%$ and $30\%$ of the total iterations to match the performance of \rlsquarex-transformer on the 11x11 and 13x13 grids, respectively. The training curves are available in Appendix \ref{sec:rl3_training_curves}.

\begin{figure}[t]
    \centering
    \includegraphics[width=\linewidth]{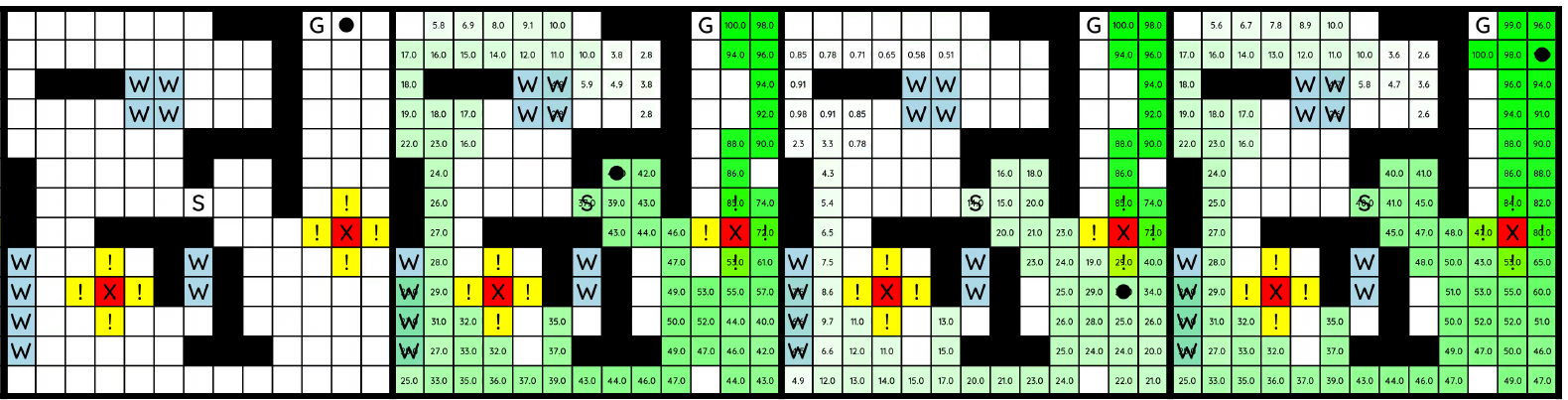}
    \caption{An \rlcube policy on a selected meta-episode visualized using a sequence of snapshots. `S' is the starting tile, `G' is the goal tile and the black circle shows the current position of the agent. Blue tiles marked `W' are wet tiles. Wet tiles always lead to the agent slipping to one of the directions orthogonal to the intended direction of movement. Entering wet tiles yield an immediate reward of -2. Yellow tiles marked `!' are warning tiles and entering them causes -10 reward. Red tiles marked `X' are fatally dangerous. Entering them ends the episode and leads to a reward of -100. Black tiles are obstacles. White tiles yield a reward of -1 to incentive the agent to reach the goal quickly. On all tiles other than wet tiles, there is a chance of slipping sideways with a probability of 0.2. The object-level state-values $v^t(s) = \text{max}_a Q^t(s,a)$, as approximated by object-level RL, is represented using shades of green (and the accompanying text), where darker shades represent higher values.}
    \label{fig:gw-example}
\end{figure}

Fig.\;\ref{fig:gw-example} shows a sequence of snapshots of a meta-episode where the trained \rlcube agent is interacting with an instance of a 13x13 grid. The first snapshot shows the agent just before reaching the goal for the first time. Prior to the first snapshot, the agent had explored many locations in the grid. The second snapshot shows the next episode just after the agent finds the goal, resulting in value estimates being updated using object-level RL for all visited states. Snapshot 3 shows the agent consequently using the $Q$-estimates to navigate to the goal presumably by choosing high-value actions. The agent also explores several new nearby states for which it does not have $Q$-estimates. Snapshot 4 shows the final $Q$-value estimates. 
A set of short videos of the GridWorlds environment, showing both \rlsquarex-transformer and \rlcube agents solving the same set of problem instances, can be found at \url{https://youtu.be/eLA_S1BQUYM}.

\noindent \textbf{Computation Overhead Considerations:} As mentioned earlier, for implementing object-level RL, we use model estimation followed by finite-horizon value-iteration to obtain $Q$-estimates. The computation overhead is negligible for Bandits (5 actions, task horizon = 1) and very little for the MDPs domain (10 states, 5 actions, task horizon 10). For 13x13 GridWorlds (up to 169 states, 5 actions, task horizon = 350), \rlcube takes approximately twice the computation time of \rlsquarex-transformer per meta-episode. However, \rlcubex-coarse requires only 10\% overhead while still outperforming \rlsquarex-transformer and retaining more than 90\% of the performance of \rlcubex. This demonstrates the utility of state abstractions in \rlcube for scaling. Finally, the meta-training sample efficiency demonstrated by \rlcube translates directly to wall-time efficiency as training is dominated by gradient computation, \emph{not} value iteration during data collection in PPO. Our implementation is available at \url{https://github.com/bhatiaabhinav/RL3}.

\section{Limitations and Conclusion}

Though it compares favorably to strong Meta-RL approaches like \rlsquarex-transformer where applicable, \rlcube does have some limitations. First, it assumes the object-level decision-making model is an MDP, which although a common assumption in the literature, may be challenged in practice. While in principle we could extend \rlcube to POMDPs using methods like point-based value iteration, this has yet to be tested empirically. Second, \rlcube relies on fast, potentially approximate methods for object-level RL, and using value iteration complicates application to problems with continuous state spaces. However, we speculate that a crude linear function approximation would suffice. Finally, inference time is slightly slower at deploy time due to running object-level RL. However, the overall training time is actually faster because of better meta-training efficiency. In fact, \rlcube could enable working with adaptation periods that are otherwise prohibitively long for many Meta-RL approaches.

To conclude, in this paper, we introduced \rlcubex, a principled hybrid approach that combines the strengths of traditional RL and Meta-RL and provides a more robust and efficient Meta-RL algorithm. We advanced intuitive and theoretical arguments regarding its suitability for Meta-RL and presented empirical evidence to validate those ideas. Specifically, we demonstrated that \rlcube holds potential to enhance long-term performance, generalization on out-of-distribution tasks and reducing meta-training time.

\appendix





\subsubsection*{Acknowledgments}
This work was supported in part by the National Science Foundation grant numbers 1954782, 2205153, and 2321786.


\bibliography{bibliography}
\bibliographystyle{rlj}

\beginSupplementaryMaterials

\section{Proofs}

\subsection{Bayes Optimality of \texorpdfstring{$Q$}{Q}-value Estimates in Bernoulli Multi-armed Bandits}\label{subsec:Bernoulli-Bandits-proof}
 
Given an instance of a Bernoulli multi-armed bandit MDP, $M_i \sim \mathcal{M}$, and trajectory data $\Upsilon_{1:T}$ up to time $T$, we would like to show that the probability $P(i| \Upsilon_{1:T})$ can be determined entirely from $Q$-estimates $Q_i^T$ and action-counts $N_i^T$, as long as the initial belief is uniform or known.

In the following proof, we represent an instance $i$ of $K$-armed Bandits as a $K$-dimensional vector of success probabilities $[p_{i1}, ..., p_{iK}]$, such that pulling arm $k$ is associated with reward distribution $P(r=1 | i,k) = p_{ik}$ and $P(r=0 | i,k) = (1-p_{ik})$.

Let the number of times arm $k$ is pulled up to time $T$ be $N_{ik}^T$, and the number of successes associated with pulling arm $k$ up to time $T$ be $q_{ik}^T$. Given that this is an MDP with just a single state and task horizon of 1, the $Q$-estimate associated with arm $k$ is just the average reward for that action, which is the ratio of successes to counts associated with that action i.e., $Q_{ik}^T = \frac{q_{ik}^T}{N_{ik}^T}$. To reduce the clutter in the notation, we will drop the superscript $T$  for the rest of the subsection. Now,
\begin{align}
    P(i | \Upsilon_{1:T}) &= \alpha P(i) \cdot P(\Upsilon_{1:T} | i)
\end{align}
\noindent where $\alpha$ is the normalization constant, $P(i)$ is the prior probability of task $i$ (which is assumed to be known beforehand), and $\Upsilon_{1:T}$ is the sequence of actions and the corresponding rewards up to time $T$. Assuming, without loss of generality, that the sequence of actions used to disambiguate tasks is a given, $P(\Upsilon_{1:T} | i)$ becomes simply the product of probabilities of reward outcomes up to time $T$, noting that the events are independent. Therefore,

\begin{align}
    P(\Upsilon_{1:T} | i) &= \prod_{k=1:K} \prod_{t=1:T} ([r_{tk}=1]p_{ik} + [r_{tk}=0](1-p_{ik})) \\
    &= \prod_{k=1:K} p_{ik}^{q_{ik}} \cdot (1 - p_{ik})^{N_{ik} - q_{ik}} \\
    &= \prod_{k=1:K} p_{ik}^{Q_{ik} N_{ik}} \cdot (1 - p_{ik})^{N_{ik} - Q_{ik}N_{ik}}
\end{align}

Putting everything together, 

\begin{align}
    P(i | \Upsilon_{1:T}) &= \alpha P(i) \cdot \prod_{k=1:K} p_{ik}^{Q_{ik} N_{ik}} \cdot (1 - p_{ik})^{N_{ik} - Q_{ik} N_{ik}}
\end{align}

This equation proves that $N_i^T$ and $Q_i^T$ are sufficient statistics to determine $P(i | \Upsilon_{1:T})$ in this domain, assuming that the prior over task distribution is known. \qed

\subsection{Non-Bayes Optimality of \texorpdfstring{$Q$}{Q}-value Estimates in Gaussian Multi-armed Bandits}\label{subsec:Gaussian-Bandits-proof}

Given an instance of a Gaussian multi-armed bandit MDP, $M_i \sim \mathcal{M}$, and trajectory data $\Upsilon_{1:T}$ up to time $t$, here we derive the closed-form expression of the probability $P(i| \Upsilon_{1:T})$ and show that it contains terms other than $Q$-estimates $Q_i^t$ and action-counts $N_i^t$.

In the following proof, we represent an instance $i$ of $K$-armed Bandits as a $2K$-dimensional vector of means and standard deviations $[\mu_{i1}, ..., \mu_{iK}, \sigma_{i1}, ..., \sigma_{iK}]$, such that pulling arm $k$ is associated with reward distribution $P(r|i,k) = \frac{1}{\sqrt{2\pi\sigma_{ik}}} \exp(\frac{r - \mu_{ik}}{\sigma_{ik}})^2$.

Let the number of times arm $k$ is pulled up to time $T$ be $N_{ik}^T$. Given this is an MDP with a single state and the task horizon is 1, the $Q$-estimate associated with arm $k$ is just the average reward for that action $\text{Avg}[r_k]$ up to time $T$. To reduce the clutter in the notation, we will drop the superscript $T$  for the rest of the subsection. As in the previous subsection, we now compute the likelihood $P(\Upsilon_{1:T} | i)$.

\begin{align}
    P(\Upsilon_{1:T} | i) &= \prod_{k=1:K} \prod_{t=1:T} \frac{1}{\sqrt{2\pi\sigma_{ik}}} \exp(\frac{r_{tk} - \mu_{ik}}{\sigma_{ik}})^2
\end{align}
Therefore, the log likelihood is
\begin{align}
    & \log P(\Upsilon_{1:T} | i) = \sum_{k=1:K} \sum_{t=1:T} \frac{(r_{tk} - \mu_{ik})^2}{\sigma_{ik}^2} - \log{(2\pi\sigma_{ik})} / 2 \\
    &= \sum_{k=1:K} N_{ik} \frac{\text{Avg}[(r_{tk} - \mu_{ik})^2]}{\sigma_{ik}^2} \nonumber \\
    & \hspace{10em} - N_{ik}\log{(2\pi\sigma_{ik})} / 2 \\
    &= \sum_{k=1:K} N_{ik} \frac{\text{Avg}[r_k^2] - 2\mu_{ik}\text{Avg}[r_k] + \mu_{ik}^2}{\sigma_{ik}^2} \nonumber \\
    & \hspace{10em} - N_{ik}\log{(2\pi\sigma_{ik})} / 2 \\
    &= \sum_{k=1:K} N_{ik} \frac{(\text{Var}[r_k] + \text{Avg}[r_k]^2) - 2\mu_{ik}\text{Avg}[r_k] + \mu_{ik}^2}{\sigma_{ik}^2} \nonumber\\
    & \hspace{10em} - N_{ik}\log{(2\pi\sigma_{ik})} / 2 \\
    &= \sum_{k=1:K} N_{ik} \frac{\text{Var}[r_k] + (Q_{ik})^2 - 2\mu_{ik}Q_{ik} + \mu_{ik}^2}{\sigma_{ik}^2} \nonumber\\
    & \hspace{10em} - N_{ik}\log{(2\pi\sigma_{ik})} / 2
\end{align}

Therefore, computing this expression requires computing the variance in rewards, $\text{Var}[r_k]$, associated with each arm up to time $T$, apart from the $Q$-estimates and action-counts. This proves that $Q$-estimates and action-counts alone are insufficient to completely determine $P(i | \Upsilon_{1:T})$ in Gaussian multi-armed bandits domain. \qed

\subsection{Object-level \texorpdfstring{$Q$}{Q}-estimates and Meta-level Values}\label{subsec:meta-value-proof}

\paragraph{\emph{Proof of Equation~\ref{eq:QVequiv}.}}
In standard Meta-RL, the POMDP state is $\bar{s}_t = [s_t, i]$, but the agent only observes the MDP state $s_t$, i.e., $\bar{\omega}_t = s_t$, while the task identity $i$ remains hidden. In \rlcubex, however, $\bar{\omega}_t$ additionally includes the vector of $Q$-estimates $Q_i^t(s_t)$ for the hidden task. Consequently, the observation function is defined as $\bar{O}\bigl(\bar{\omega}_t \mid \bar{s}_t, a_{t-1}\bigr) = \Pr\bigl(\bar{\omega}_t = [s_t, Q_i^t(s_t)] \mid [s_t, i], a_{t-1}\bigr)$, encoding the probability that the observed $Q$-values are consistent with task $i$. Because the observation model appears in the belief update rule (see Section~\ref{sec:pomdp}), and because $Q$-estimates provide strong evidence for task identification (Appendix~\ref{sec:analysis}), robust recovery of the task belief is possible even with a noisy prior. Once the $Q$-estimates converge, consider the following two cases:

\textbf{Case 1 (unique $Q$-signature).} If the observed $Q$-values are unique to MDP $M_i$, the belief distribution rapidly collapses to zero probability for all tasks $j \neq i$, yielding $\bar{V}^*(\bar{b}) = \max_{a \in A} Q_i(s,a)$.

\textbf{Case 2 (non-unique $Q$-signature).} If several tasks share identical $Q$-values, the belief concentrates only on this subset of tasks. Since all remaining tasks induce the same $Q$-values, $\bar{V}^*(\bar{b})$ still simplifies to $\max_{a \in A} Q_i(s,a)$, where $i$ denotes any of the tasks with non-zero belief.

This establishes Equation~\ref{eq:QVequiv}. In the limit, after all state-action pairs have been explored, the task identity can be inferred exactly, reducing the meta-level value function to the optimal object-level value. However, the critical insight of Equation~\ref{eq:QVequiv} is that \rlcube achieves this equivalence directly through $Q$-based observations, without relying on the full experience stream. \qed

\paragraph{\emph{Proof of Equation~\ref{eq:bamdp}.}}
The optimal meta-level value function in belief space satisfies the Bellman equation:
\begin{equation}
    \bar{V}^*(\bar{b}) = \max_{a \in A} \Big [ \sum_{\bar{s} \in \bar{S}} \bar{b}(\bar{s}) \bar{R}(\bar{s},a) + \gamma \sum_{\bar{\omega} \in \bar{\Omega}} \Pr(\bar{\omega}|\bar{b},a) \bar{V}^*(\bar{b}') \Big ].
\end{equation}
Because the only hidden variable in $\bar{s} = [s,i]$ is the task identity $i$, this expression reduces to
\begin{equation}
    \bar{V}^*(\bar{b}) = \max_{a \in A} \Big [ \sum_{M_i \in \mathcal{M}} \bar{b}(i) R_i(s,a) + \gamma \sum_{\bar{\omega} \in \bar{\Omega}} \Pr(\bar{\omega} |\bar{b},a) \bar{V}^*(\bar{b}') \Big ],
\end{equation}

where the observation likelihood is computed as $\Pr(\bar \omega | \bar b, a) = \sum_{\bar s'} \bar O(\bar \omega | \bar s', a) \sum_{\bar s} \Pr(\bar s' | \bar s, a) \cdot \bar b(\bar s)$. \qed

\section{Architecture}\label{sec:architecture}

\subsection{RL\texorpdfstring{$^2$}{2}}

Our modified implementation of \rlsquare uses transformer decoders~\citep{vaswani2017attention} instead of RNNs to map trajectories to action probabilities and meta-values, in the actor and the critic, respectively, and uses PPO instead of TRPO for outer RL. The decoder architecture is similar to~\citep{vaswani2017attention}, with 2 layers of masked multi-headed attention. However, we use learned position embeddings instead of sinusoidal, followed by layer normalization. Our overall setup is similar to~\citep{esslinger2022deep}.

For each meta-episode of interactions with an MDP $M_i$, the actor and the critic transformers look at the entire history of experiences up to time $t$ and output the corresponding action probabilities $\pi_1 ... \pi_t$ and meta-values $\bar V_1 ... \bar V_t$, respectively. An experience input to the transformer at time $t$ consists of the previous action $a_{t-1}$, the latest reward $r_{t-1}$, the current state $s_t$, episode time step $t_\tau$, and the meta-episode time step $t$, all of which are normalized to be in the range $[0, 1]$. In order to reduce inference complexity, say at time step $t$, we append $t$ new attention scores (corresponding to experience input $t$ w.r.t.~the previous $t-1$ experience inputs) to a previously cached $(t-1) \times (t-1)$ attention matrix, instead of recomputing the entire $t \times t$ attention matrix. This caching mechanism is implemented for each attention head and reduces the inference complexity at time $t$ from $\mathcal{O}(t^2)$ to $\mathcal{O}(t)$.

\subsection{RL\texorpdfstring{$^3$}{3}}

The input of the transformer in \rlcube includes a vector of $Q$ estimates (in practice, they are supplied as the vector of advantage estimates ($Q - \text{max}_aQ$) along with the value function ($\text{max}_aQ$) separately) and a vector of action counts at each step $t$ for the corresponding state. As mentioned in Section \ref{sec:impl}, this is implemented in our code simply by converting MDPs in the problem set to VAMDPs using a wrapper and running our implementation of \rlsquare thereafter. The pseudocode is shown in the algorithm \ref{alg:vamdp}. The Markov version of \rlcube uses a dense neural network, with two hidden layers of 64 nodes each, with the ReLU activation function.

\begin{algorithm}[t]
    \caption{Value-Augmenting Wrapper for Discrete MDPs}
    \begin{algorithmic}
        \Procedure{ResetMDP}{vamdp}
            \State vamdp.$t \gets 0$; vamdp.$t_\tau \gets 0$
            \State vamdp.$N[s,a] \gets 0$; vamdp.$Q[s,a] \gets 0 \quad \forall s \in S, a \in A$
            \State vamdp.rl $\gets \textsc{InitRL}()$
            \State $s = \textsc{ResetMDP}$(vamdp.mdp)
            \State \Return $\textsc{OneHot}(s) \cdot Q[s] \cdot N[s]$
        \EndProcedure
        \Procedure{StepMDP}{vamdp, $a$}
            \State $s \gets $ mdp.$s$
            \State $r, s' \gets \textsc{StepMDP}$(vamdp.mdp, $a$)
            \State $d \gets \textsc{Terminated}$(vamdp.mdp)
            \State vamdp.$t$, vamdp.$N[s,a]$, vamdp.$t_\tau \gets \hspace{1mm} \mathrel{+}= 1$
            \State vamdp.$Q \gets \textsc{UpdateRL}$(vamdp.rl, $s, a, r, s', d$) 
            \If {$d$ \textbf{ or } vamdp.$t_\tau \ge \texttt{task\_horizon}$}
                \State vamdp.$t_\tau \gets 0$
                \State $s' \gets \textsc{ResetMDP}$(vamdp.mdp)
            \EndIf
            \State \Return $r, $ $\textsc{OneHot}(s') \cdot Q[s'] \cdot N[s']$ \Comment{Concatenate state, $Q$-estimates and action counts}
        \EndProcedure
        \Procedure{Terminated}{vamdp}
            \State \Return vamdp.$t \ge H$
        \EndProcedure
    \end{algorithmic}
    \label{alg:vamdp}
\end{algorithm}

For object-level RL, we use model estimation followed by value iteration (with discount factor $\gamma=1$) to obtain $Q$-estimates.
The transition probabilities and the mean rewards are estimated using maximum likelihood estimation (MLE), with Laplace smoothing (coefficient = 0.1) for transition probabilities estimation. For unseen actions, rewards are assumed to be zero, and transitions equally likely to other states.
States are added to the model incrementally when they are visited, so that value iteration does not compute values for unvisited states. Moreover, value iteration is carried out only for iterations equal to the task horizon (which is 1, 10, 250, 350 for Bandits, MDPs, 11x11 GridWorlds, 13x13 GridWorlds domains, respectively), unless the maximum Bellman error drops below 0.01.

\subsection{RL\texorpdfstring{$^3$}{3}-coarse}\label{sec:rl3appendix}

During model estimation in \rlcubex-coarse, concrete states in the underlying MDP are incrementally clustered into abstract states as they are visited. When a new concrete state is encountered, its abstract state ID is set to that of a previously visited state within a `clustering radius', unless that previous state is already part of a full cluster (determined by a maximum `cluster size' parameter). If multiple visited states satisfy the criteria, the ID of the closet one is chosen. If none of the visited states that satisfy the criteria, then the new state is assigned a new abstract state ID, increasing the number of abstract states in the model. It is worth noting that this method of deriving abstractions does not take advantage of any structure in the underlying domain. However, this simplicity makes it general purpose, efficient, and impartial, while still leading to excellent performance. For our GridWorlds domain, we chose a cluster size of 2 and a clustering radius such that only non-diagonal adjacent states are clustered (Manhattan radius of 1). 

The mechanism for learning the transition function and the reward function in the abstract MDP is the same as before. For estimating $Q$-values for a given concrete state, value iteration is carried out on the abstract MDP and the $Q$-estimates of the corresponding abstract state are returned.

\begin{figure}
    \centering
    \begin{subfigure}{0.45\linewidth}
        \includegraphics[width=\linewidth]{figures/training_curves/mdps-128.pdf}
        \caption{$H=128$}
    \end{subfigure}
    \begin{subfigure}{0.45\linewidth}
        \includegraphics[width=\linewidth]{figures/training_curves/mdps-256.pdf}
        \caption{$H=256$}
    \end{subfigure}
    \begin{subfigure}{0.45\linewidth}
        \includegraphics[width=\linewidth]{figures/training_curves/mdps-512.pdf}
        \caption{$H=512$}
    \end{subfigure}
    \begin{subfigure}{0.45\linewidth}
        \includegraphics[width=\linewidth]{figures/training_curves/mdps-1024.pdf}
        \caption{$H=1024$}
    \end{subfigure}
    \begin{subfigure}{0.45\linewidth}
        \includegraphics[width=\linewidth]{figures/training_curves/mdps-1500.pdf}
        \caption{$H=1500$}
    \end{subfigure}
    \begin{subfigure}{0.45\linewidth}
        \includegraphics[width=\linewidth]{figures/training_curves/mdps-2048.pdf}
        \caption{$H=2048$}
    \end{subfigure}
    \caption{Average meta-episode return vs PPO iterations for MDPs domain for different interaction budgets.}
    \label{fig:training_curves:mdps}
\end{figure}

\begin{figure}
    \centering
    \includegraphics[width=0.45\linewidth]{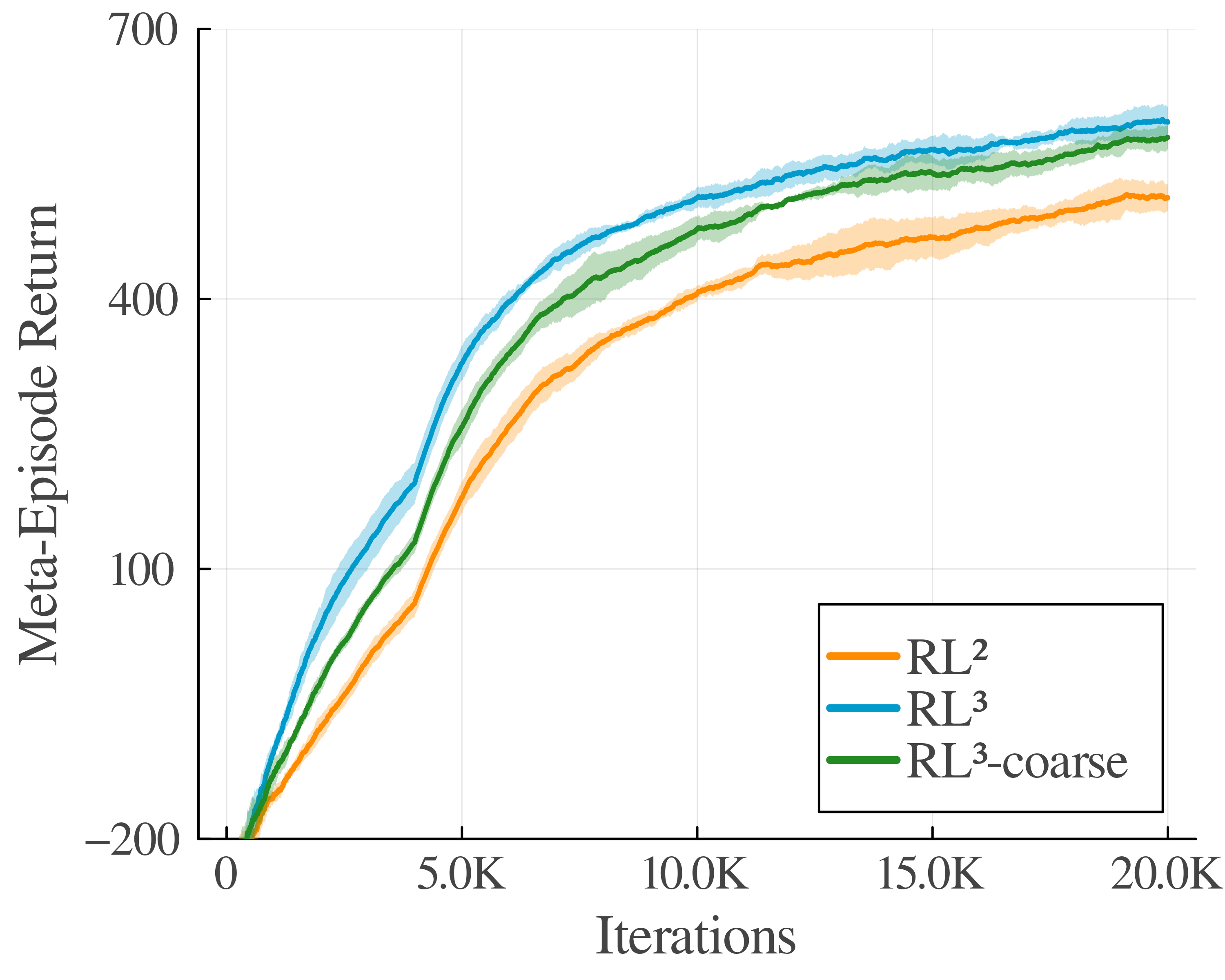}
    \includegraphics[width=0.45\linewidth]{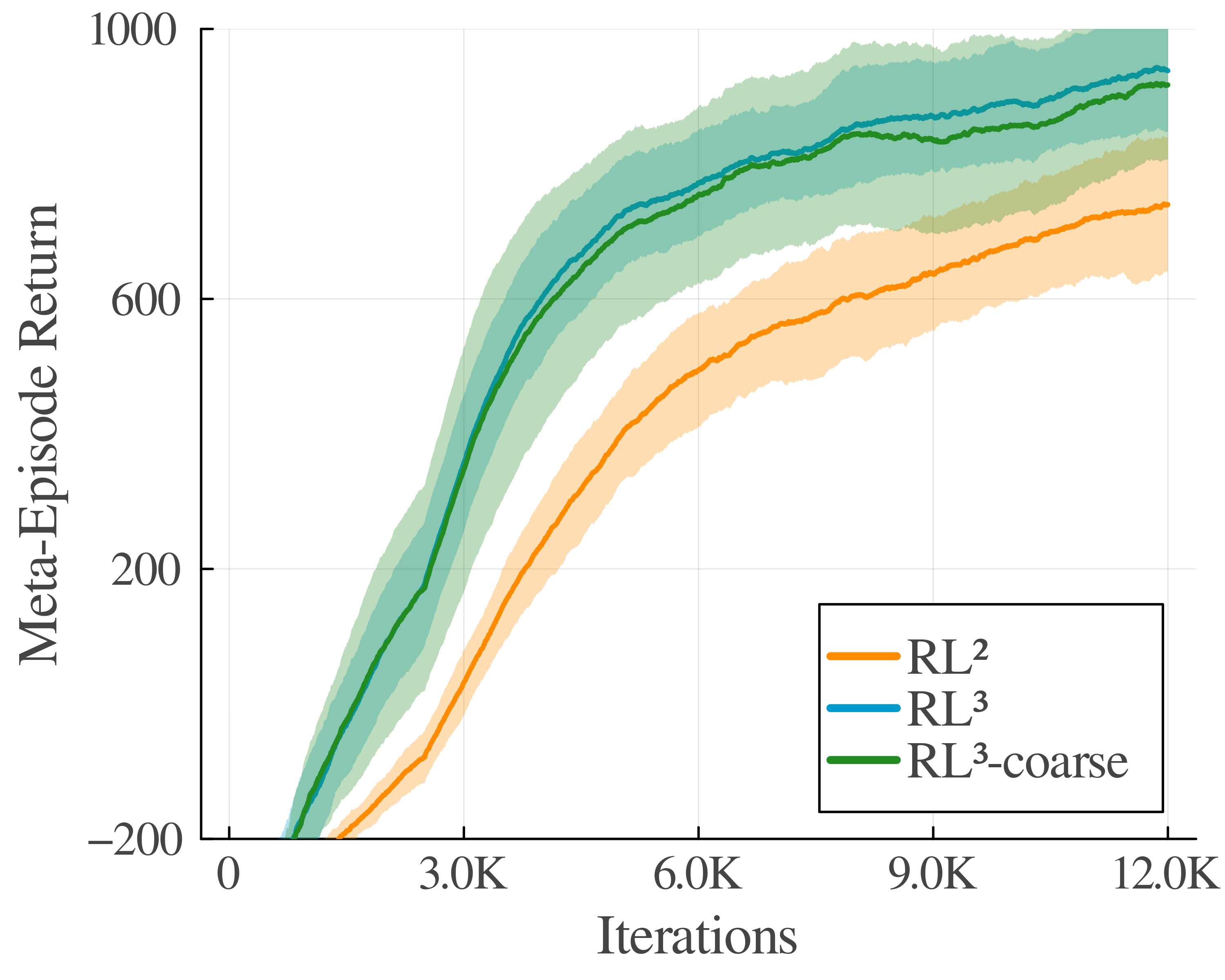}
    \caption{Average meta-episode return vs PPO iterations for GridWorlds 11x11 (\textit{left}) and 13x13 $(\textit{right})$.}
    \label{fig:training_curves:gridworlds}
\end{figure}

\section{Training}\label{sec:rl3_training_curves}

Figs.~\ref{fig:training_curves:mdps}, and \ref{fig:training_curves:gridworlds} show the training curves for MDPs, and GridWorlds environments, respectively, across 3 random seeds. The results in the main text correspond to the median model. We ran the experiments on Nvidia GeForce RTX 2080 Ti GPUs for context length $\leq 256$ which took approximately 12-24 hours, and on Nvidia A100 GPUs for higher context lengths, which took 1-2 days.

\section{Additional Analysis}\label{sec:analysis}

In this section, we show that $Q$-estimates, though imperfect, produce reasonable signals for task identification. Here, we test this claim thoroughly with 3 analyses.

\subsection{Requirements for a Unique \texorpdfstring{$Q^*$}{Q*}-Function} 

Throughout, we assume fixed state space and action space. Below, we show that if the transition function is fixed, then two $Q^*$-tables will be identical if and only if both reward functions are also equal. First, we show that identical $Q^*$ functions imply identical reward functions. Given the Bellman equations,
\begin{small}
\begin{align}
    Q^*_1(s, a) &= R_1(s, a) + \gamma \sum_{s'} T(s, a, s') \text{max}_{a'} Q^*_1(s', a') \label{eq:bellmanQ1}\\
    Q^*_2(s, a) &= R_2(s, a) + \gamma \sum_{s'} T(s, a, s') \text{max}_{a'} Q^*_2(s', a') \label{eq:bellmanQ2}
\end{align}
\end{small}

\noindent Substituting $Q^*_2 = Q^*_1$ in Equation \eqref{eq:bellmanQ2}, we get
\begin{small}
\begin{align}
    Q^*_1(s, a) &= R_2(s, a) + \gamma \sum_{s'} T(s, a, s') \text{max}_{a'} Q^*_1(s', a') \label{eq:bellmanQ2sub}
\end{align}
\end{small}

\noindent Subtracting Equation \eqref{eq:bellmanQ1} from Equation \eqref{eq:bellmanQ2sub}, we get $R_1(s, a) = R_2(s, a)$. Thus, $(Q^*_1 = Q^*_2) \land (T_1 = T_2) \implies (R_1 = R_2)$.

Now, if two MDPs have the same reward and transition function, they are the same MDP and will have the same optimal value function. So, $(R_1 = R_2) \land (T_1 = T_2) \implies (Q^*_1 = Q^*_2)$.

Since encountering similar $Q^*$-tables is thus dependent on both transitions and rewards `balancing' each other, the question is then for practitioners: How likely are we to get many MDPs that all appear to have very similar $Q^*$-tables?

\subsection{Empirical Test using Max Norm}

Given an MDP with 3 states and 2 actions, we want to find the probability that $||Q^*_1 - Q^*_2||_\infty < \delta$, where $Q^*_1$ and $Q^*_2$ are 6-entry (3 states $\times$ 2 actions) $Q^*$-tables. The transition and reward functions are drawn from distributions parameterized by $\alpha$ and $\beta$, respectively. Transition probabilities are drawn from a Dirichlet distribution, Dir($\alpha$), and rewards are sampled from a normal distribution, $\mathcal{N}(1, \beta)$. In total, we ran 3 combinations of $\alpha$ and $\beta$, each with 50,000 MDPs, a task horizon of 10, and $\delta=0.1$. To get the final probability, we test all $((50,000-1)^2)/2$ non-duplicate pairs and count the number of max norms less than $\delta$.

{\bf Results:} For $\alpha=1.0$, $\beta=1.0$, we found the probability of a given pair of MDPs having duplicate $Q^*$-table to be $\epsilon = 2.6 \times 10^{-9}$. For $\alpha=0.1$, $\beta=1.0$, which is a more deterministic setting, we found $\epsilon=4.6\times10^{-9}$. Further, with $\alpha=0.1$, $\beta=0.5$, where rewards are more closely distributed, we found $\epsilon=1.1\times10^{-7}$. Overall, we can see that even for a set of very small MDPs, the probability of numerically mistaking one $Q^*$-table for another is vanishingly small.

\subsection{Predicting Task Families}

The near uniqueness of $Q^*$-functions is encouraging, but max norm is not a very sophisticated metric. Here, we test whether a very simple multi-class classifier (1 hidden layer of 64 nodes), can accurately identify individual tasks based on their $Q$-\emph{estimates}. Moreover, we track how the classification accuracy improves as a function of the number of steps taken within the MDP as the estimates improve. In this experiment, the same random policy is executed in each MDP for 50 time steps. As before, our MDPs have 3 states and 2 actions. 

\begin{figure}
    \centering
    \includegraphics[width=0.45\linewidth]{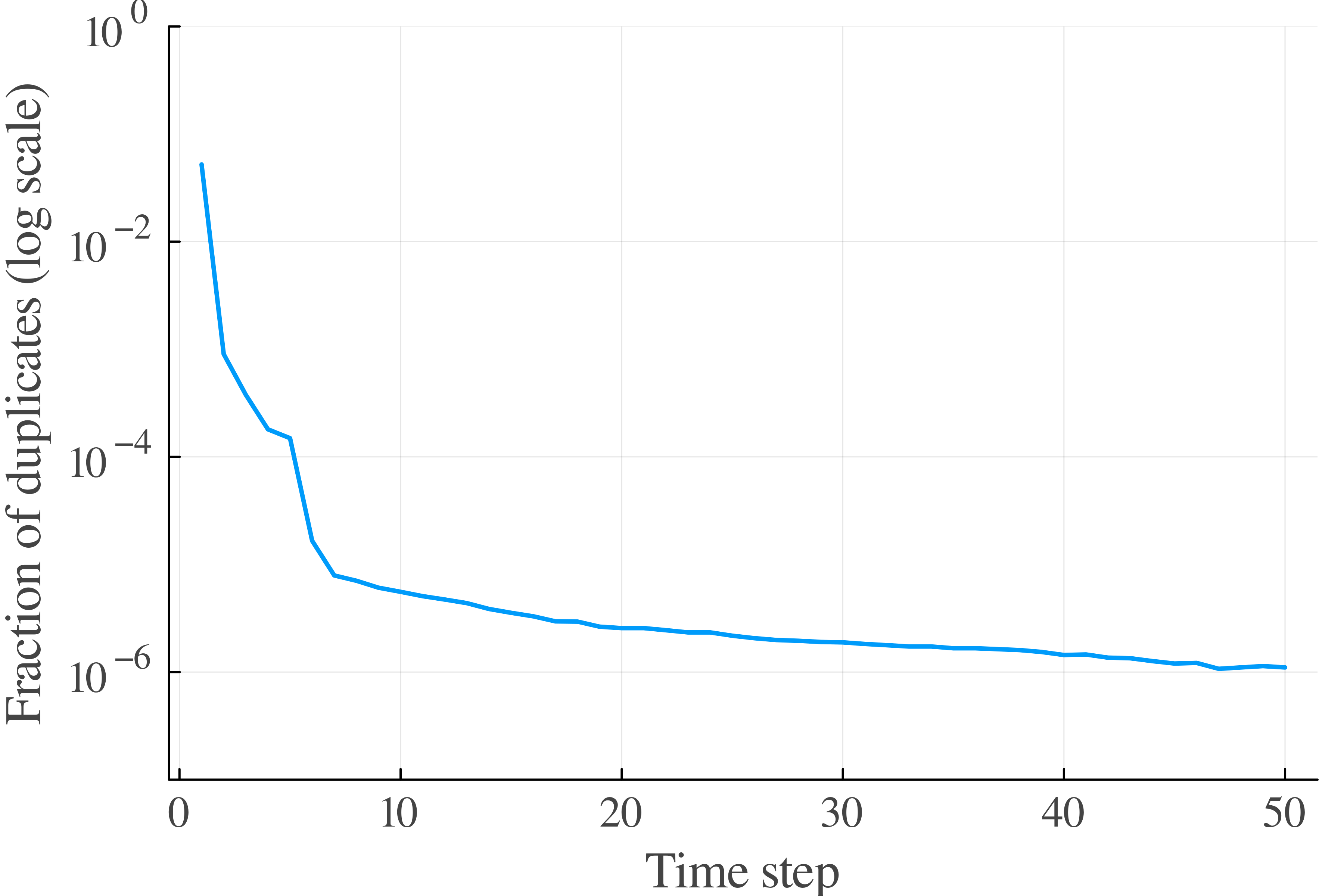}
    \includegraphics[width=0.45\linewidth]{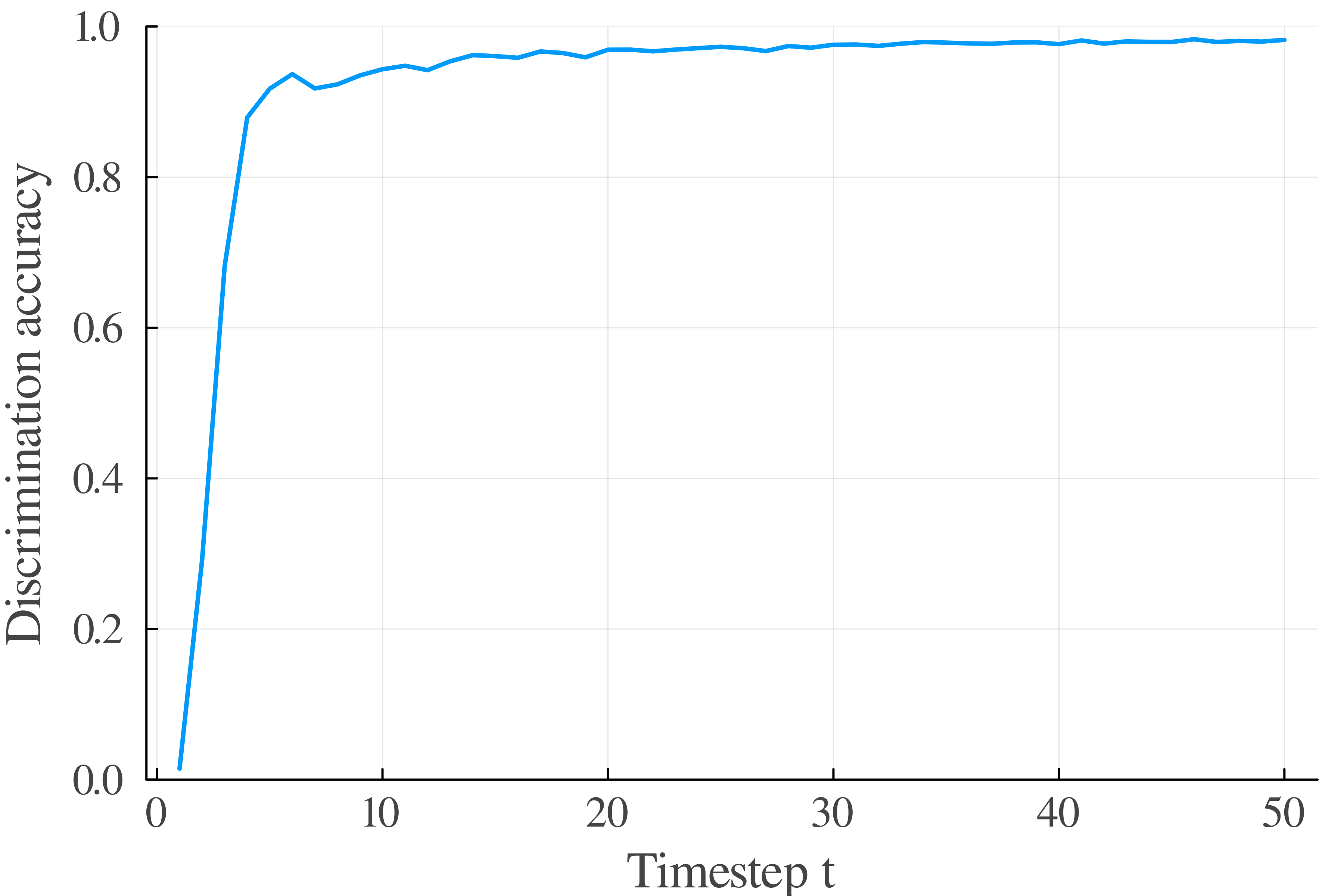}
    \caption{The task-identification power of $Q$-estimates. \textit{Left}: Fraction of $\delta$-duplicates, with $\delta=0.1$, as a function of time steps in a set of 5,000 random MDPs. \textit{Right}: Accuracy of a simple multi-class classifier in predicting task ID given $Q$-table estimates, as function of time step. Both figures are generated using the same policy.}
    \label{fig:qdiscrimination}
\end{figure}

We instantiate 10,000 MDPs whose transition and reward functions are drawn from the same distribution as before: transitions from a Dirichlet distribution with $\alpha=0.1$ and rewards sampled from a normal distribution $N(1, 0.5)$. Thus, this is a classification problem with 10,000 classes. \emph{A priori}, this exercise seems relatively difficult given the number of tasks and the parameters chosen for the distributions. Fig.\;\ref{fig:qdiscrimination} shows a compelling result given the simplicity of the model and the relative difficulty of the classification problem. Clearly, $Q$-estimates, even those built from only 20 experiences, provide a high signal-to-noise ratio w.r.t.~ task identification. And this is for a \emph{random} policy. In principle, the Meta-RL agent could follow a much more deliberate policy that actively disambiguates trajectories such that the $Q$-estimates evolve in a way that leads to faster or more reliable discrimination.

\section{Domain Descriptions}\label{sec:domains}

\subsection{Bernoulli Multi-Armed Bandits}

We use the same setup described by~\citet{duan2016rl}. At the beginning of each meta-episode, the success probability corresponding to each arm is sampled from a uniform distribution $\mathcal U(0, 1)$. To test OOD generalization, we sample success probabilities from $\mathcal{N}(0.5, 0.5)$

\subsection{Random MDPs}

We use the same setup described by~\citet{duan2016rl}. The MDPs have 10 states and 5 actions. For each meta-episode, the mean rewards $R(s, a)$ and transition probabilities $T(s, a, s')$ are initialized from a normal distribution ($\mathcal N(1, 1)$) and a flat Dirichlet distribution ($\alpha=1$), respectively. Moreover, when an action $a$ is performed in state $s$, a reward is sampled from $\mathcal N(R(s, a), 1)$. To test OOD generalization, the transition probabilities are initialized with Dirichlet $\alpha=0.25$.

Each episode begins at state $s=1$ and ends after \texttt{task\_horizon}~=~10 time steps.

\subsection{GridWorlds}

A set of navigation tasks in a 2D grid environment. We experiment with 11x11 (121 states) and 13x13 (169 states) grids. The agent always starts in the center of the grid and needs to navigate through obstacles to a single goal location. The goal location is always at a minimum of \texttt{min goal manhat} Manhattan distance from the starting tile.
The grid also contains slippery wet tiles, fatally dangerous tiles and warning tiles surrounding the latter. There are \texttt{num obstacle sets} set of obstacles, and each obstacle set spans \texttt{obstacle set len} tiles, in either horizontal or vertical configuration. There are \texttt{num water sets} set of wet regions and each wet region always spans \texttt{water set length}, in either a horizontal or vertical configuration. Entering wet tiles yields an immediate reward of -2. There are \texttt{num dangers} danger tiles and entering them ends the episode and leads to a reward of -100. Warning tiles always occur as a set of 4 tiles non-diagonally surrounding the corresponding danger tiles. Entering warning tiles causes -10 reward. Normal tiles yield a reward of -1 to incentivize the agent to reach the goal quickly. On all tiles, there is a chance of slipping sideways with a probability of 0.2, except for wet tiles, where the probability of slipping sideways is 1.

The parameters for our canonical 11x11 and 13x13 GridWorlds are as follows: \allowbreak
\texttt{num obstacle sets}~=~11, \allowbreak \texttt{obstacle set len}~=~3, \allowbreak \texttt{num water sets}~=~5, \allowbreak \texttt{water set length}~=~2, \allowbreak \texttt{num dangers}~=~2, and \allowbreak \texttt{min goal manhat}~=~8. The parameters for the OOD variations are largely the same and the differences are as follows. For \textsc{Deterministic} variation, the slip probability on non-wet tiles is 0. For \textsc{Dense} variation, \texttt{obstacle set len} is increased to 4. For \textsc{Watery} variation, \texttt{num water sets} is increased to 8. For \textsc{Dangerous} variation, \texttt{num dangers} is increased to 4. For \textsc{Corner} variation, \texttt{min goal manhat} is set to 12, so that the goal is placed on one of the corners of the grid.

There is no fixed task horizon for this domain. An episode ends when the agent reaches the goal or encounters a danger tile. In principle, an episode can last through the entire meta-episode if a terminal state is not reached.

When a new grid is initialized at the beginning of each meta-episode, we ensure that the optimal, non-discounted return within a fixed horizon of 100 steps is between 50 and 100. This is to ensure that the grid both has a solution and the solution is not trivial.

\begin{table}
    \caption{\rlsquare/\rlcube Hyperparameters}
    \label{tab:hyperparams}
    \centering
    \begin{tabular}{l l}
    \textbf{Hyperparameter}  & \textbf{Value} \\
    \hline \\
    Learning Rate (Actor and Critic) & 0.0003 (Bandits, MDPs) \\
    & 0.0002 (GridWorlds) \\
    Adam $\beta1, \beta2, \epsilon$ &  $0.9, 0.999, 10^{-7}$ \\
    Weight Decay (Critic Only) & $10^{-2}$ \\
    Batch size & 32768 \\
    Rollout Length & Interaction Budget ($H$) \\
    Number of Parallel Environments & Batch Size $\div$ $H$ \\
    Minibatch Size & 4096 \\
    Entropy Regularization Coefficient & 0.1 with decay (MDPs) \\
    & 0.04 (GridWorlds) \\
    & 0.01 (Bandits) \\
    PPO Iterations & See training curves \\
    Epochs Per Iteration & 8 \\
    Max KL Per Iteration & 0.01  \\
    PPO Clip $\epsilon$ & 0.2 \\
    GAE $\lambda$ & 0.3 \\
    Discount Factor $\gamma$ & 0.99 \\
    Decoder Layers & 2 \\
    Attention Heads & 4 \\
    Activation Function & GELU \\
    Decoder Size ($d\_{model}$) & 64 \\
    \end{tabular}
\end{table}

\end{document}